\documentclass[journal]{IEEEtran}
\ifCLASSINFOpdf
\else
\fi

\usepackage{amsfonts}
\usepackage{amssymb}
\usepackage{mathrsfs}

\usepackage[cmex10]{amsmath}
\usepackage{times}
\usepackage{epsfig}
\usepackage{graphicx}
\usepackage{amsmath}
\usepackage{amssymb}
\usepackage{booktabs}
\usepackage[numbers, sort&compress]{natbib}
\usepackage{amsthm}
\usepackage{amssymb,amsmath,amsthm,enumitem}
\usepackage{mathrsfs}

\newtheorem{thm}{Theorem}
\usepackage[T1]{fontenc}

\usepackage[table,xcdraw]{xcolor}
\usepackage{subcaption}
\usepackage{floatrow}
\definecolor{Mycolor2}{HTML}{00F9DE}
\usepackage{tikz}
\usepackage{rotating}

\usepackage[export]{adjustbox}
\usepackage{amssymb}
\usepackage{pifont}
\newcommand{\cmark}{\ding{51}}%
\newcommand{\xmark}{\ding{55}}%
\begin{document}
%

\title{CLiSA: A Hierarchical Hybrid Transformer Model using Orthogonal Cross Attention for Satellite Image Cloud Segmentation}

%
%
%

\author{Subhajit Paul, Ashutosh Gupta
\thanks{High Resolution Data Processing Division, Signal and Image Processing Area, Space Applications Centre, ISRO, Ahmedabad, India-380015}
}

%
%

\markboth{}{}
%



\maketitle

\begin{abstract}
Clouds in optical satellite images are a major concern since their presence hinders the ability to carry accurate analysis as well as processing. Presence of clouds also affects the image tasking schedule and results in wastage of valuable storage space on ground as well as space-based systems. Due to these reasons, deriving accurate cloud masks from optical remote-sensing images is an important task. Traditional methods such as threshold-based, spatial filtering for cloud detection in satellite images suffer from lack of accuracy. In recent years, deep learning algorithms have emerged as a promising approach to solve image segmentation problems as it allows pixel-level classification and semantic-level segmentation. In this paper, we introduce a deep-learning model based on hybrid transformer architecture for effective cloud mask generation named CLiSA - Cloud segmentation via Lipschitz Stable Attention network. In this context, we propose an concept of orthogonal self-attention combined with hierarchical cross attention model, and we validate its lipschitz stability theoretically and empirically. We design the whole setup under adversarial setting in presence of Lovász-Softmax loss. We demonstrate both qualitative and quantitative outcomes for multiple satellite image datasets including Landsat-8, Sentinel-2, and Cartosat-2s. Performing comparative study we show that our model performs preferably against other state-of-the-art methods and also provides better generalization in precise cloud extraction from satellite multi-spectral (MX) images. We also showcase different ablation studies to endorse our choices corresponding to different architectural elements and objective functions. 
\end{abstract}
\begin{IEEEkeywords}
Cloud segmentation, cross attention, hybrid transformer, Lipschitz stability, orthogonal attention.
\end{IEEEkeywords}
%
\IEEEpeerreviewmaketitle
\section{Introduction}
%
%
%
%
\IEEEPARstart{C}{}louds in satellite images are certainly of immense interest to the meteorological community, but for optical remote-sensing applications, the presence of clouds poses several challenges. Data affected by large cloud cover is difficult to process and analyze. Geophysical parameters such as radiance and brightness temperature derived from cloud-affected optical images do not match well with reference or ground-truth values. Data processing steps such as ortho-rectification and image filtering also need to be aware of cloudy pixels to avoid outliers and artifacts in processed data. The presence of clouds also affects satellite tasking required to cover a specific area of interest at a given time window. A large amount of cloudy data is often acquired, transmitted, and stored in data processing systems essentially wasting valuable bandwidth and storage resources. Given these challenges, a preprocessing model that generates precise pixel-wise cloud masks can be very useful to avoid artifacts in images and improve the overall usability of remote sensing data.

\begin{figure}[!tb]
\centering
\begin{subfigure}{0.3\linewidth}
    \caption*{Input Image} 
    \adjustbox{width=\linewidth, frame=0pt 1pt}{\includegraphics{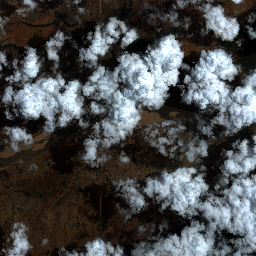}}\\
    \adjustbox{width=\linewidth, frame=0pt 1pt}{\includegraphics{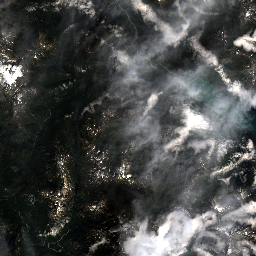}}\\
    \adjustbox{width=\linewidth, frame=0pt 1pt}{\includegraphics{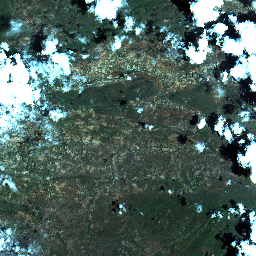}}
\end{subfigure}
\begin{subfigure}{0.3\linewidth}
    \caption*{Ground Truth Mask}
    \adjustbox{width=\linewidth, frame=0pt 1pt}{\includegraphics{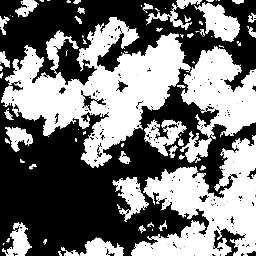}}\\
    \adjustbox{width=\linewidth, frame=0pt 1pt}{\includegraphics{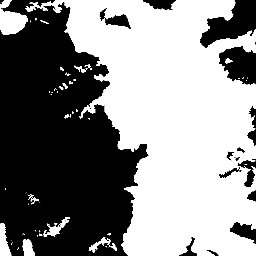}}\\
    \adjustbox{width=\linewidth, frame=0pt 1pt}{\includegraphics{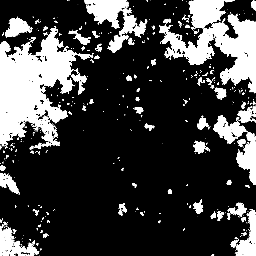}}
\end{subfigure}
\begin{subfigure}{0.3\linewidth}
    \caption*{Predicted Mask}
    \adjustbox{width=\linewidth, frame=0pt 1pt}{\includegraphics{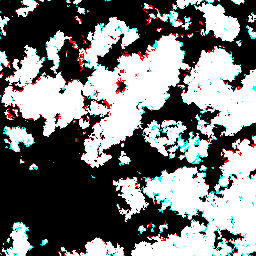}}\\
    \adjustbox{width=\linewidth, frame=0pt 1pt}{\includegraphics{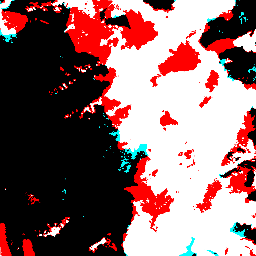}}\\
    \adjustbox{width=\linewidth, frame=0pt 1pt}{\includegraphics{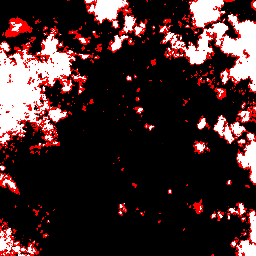}}
\end{subfigure}
 \tikz \fill [red] rectangle (1, 0.2);  Omission error \quad
\tikz \fill [Mycolor2] rectangle (1, 0.2);  Commission error
\caption{Sample results for cloud mask generation for different satellite images: Cartosat-2s, Landsat-8 and Sentinel-2 from top to bottom.}
\label{fig:fig1_results}
\end{figure}

At first, cloud mask generation may appear to be a simple binary classification problem which was traditionally performed by thresholding, spatial filtering, texture analysis, and feature-wise classification. However, these methods suffer from several limitations and complexity in generating accurate cloud masks. This is mainly because clouds exhibit large variations in thickness, type, and scale which make it difficult to put them under a single `cloudy' class during the classification process and, also, the texture and color of ground features such as snow, ice, white sand, etc. often match that of clouds, further affecting the overall accuracy of these cloud masking techniques. Recent advances in deep learning networks and optimization techniques show compelling results for cloud segmentation tasks. However, they still exhibit errors in classification for said complex scenarios, and, to add to this complexity, they are not designed to perform cross-domain images acquired by the sensors having different spatial and spectral resolutions, dynamic range, noise level, and spectral band definitions. A good cloud masking model should work efficiently in the presence of confusing ground features as well as have better generalization capability to act on images with different cloud types, and spatial and spectral resolution. In this study, we propose a novel cloud segmentation framework that works robustly for intricate cases and also showcases better generalization for cross-domain inference with minimal computation. Our key contributions can be summarized as
\begin{enumerate}
    \item We propose a novel hybrid transformer architecture, namely CLiSA based on dual orthogonal self-attention and hierarchical cross channel attention.
    \item We provide theoretical and empirical justification for the Lipschitz stability of our proposed attention module and compare it with other popular attention models.
    \item We devise our objective function in an adversarial setup along with introducing Lovász-Softmax loss due to its smoothness property compared to Jaccard loss.
    \item Finally, we perform experiments on various satellite image dataets with different band combinations and also carry out ablation studies to study the impact of different configuration choices.
\end{enumerate}
The rest of the paper is organized as follows. In section \ref{sec2} we briefly discuss prior work relevant to our study. We discuss detailed methodology along with describing the overall architecture and loss formulation in section \ref{method}. Dataset selection and implementation details are provided in section \ref{exp}. Detailed experimental results and analysis along with a comparative study is carried out in section \ref{res} followed by an ablation study in section \ref{abl}. Finally, we draw our concluding remarks in section \ref{conc}.
\section{Related Work}\label{sec2}
Cloud detection from satellite images has been an active research area in the domain of remote sensing. Traditionally, cloud mask was generated from satellite images via thresholding where it sets a specific threshold on intensity or color to separate clouds from the background \cite{4}, and, region and contour-based image segmentation where it groups pixels with similar attributes or extract boundaries to identify cloud regions \cite{5}. However, both of these methods suffer from variations in lighting conditions and complex cloud formations. To resolve these issues, one of the first pioneering and more generic cloud detection methods was Fmask \cite{26} and later, its updated versions \cite{6}. This algorithm uses spectral information and correlation between different bands of MX imagery to segment every pixel to a specific class. Although this works preferably in most situations, the spectral analysis between different bands is affected by factors like atmospheric conditions, the presence of different noise, and limited spectral bands. This not only restricts the overall performance but also becomes impractical to use for other satellite images.\par
After the emergence of deep learning based algorithms, their efficacy in the domain of semantic segmentation became notable. Due to their growing accomplishment in this domain, several fully convolutional neural networks (FCN) have been developed for cloud detection tasks. Francis \textit{et al.} introduced CloudFCN \cite{4009} as one of the leading-edge models for pixel-level cloud segmentation. They developed this model based on U-Net \cite{1} with the addition of inception modules between skip connections. Yang \textit{et al.} \cite{19} proposed a convolutional neural network (CNN) based cloud segmentation module, namely CDNetV1 which is incorporated with a built-in boundary refinement module for thumbnail images of ZY-3 satellite. Later, they proposed an updated version of the network, CDNetV2 \cite{20} which was equipped with a new feature fusion and information guidance module for precise cloud mask extraction. They designed this module specifically to work under complex conditions like scenarios of snow-cloud coexistence. Jeppesen
\textit{et al.} \cite{18} introduced the RS-Net model which was a direct implication of U-Net \cite{1} for cloud segmentation. Their main goal was to generate a cloud mask for Landsat-8 images, and for that, they trained their model with a manually generated cloud mask as well as a mask generated using the Fmask algorithm. The authors show that the results obtained from the model trained with outputs from the Fmask algorithm outperform the outcomes obtained directly from Fmask. In another work, recently Jiao \textit{et al.} \cite{4010}, has proposed an updated version of previous works, Refined UNet V4 (RefUNetV4), where they have used U-Net to coarsely locate cloudy regions followed by an upgraded conditional random field (CRF) for refining edges. By experimenting on the Landsat-8 dataset, they can extract edge precise cloud masks from satellite images.\par
After the introduction of the transformer for vision activities \cite{4011}, multiple transformer-based architectures have started to come into highlight for different vision activities. In the field of image segmentation, one such pioneering study is SegFormer \cite{4012}. These models excel at learning rich and abstract feature representation from the input data to capture complex patterns and crucial semantic information. They also aggregate information for all image regions using the self-attention mechanism to capture relationships between spatially distant but semantically related regions. However, as the transformer models majorly excel in global feature extraction, they cannot exact local dense features compared to CNN. To leverage the benefit of both the CNN and transformer model, recently, several hybrid transformer models have recently come into the picture. UTNet \cite{4015} is such a model in the medical image segmentation domain with superior performance against benchmark approaches. Despite the advantages, there is no direct implication of this concept in the cloud segmentation domain. Hence, in this study, we propose a hybrid transformer model based on a Lipschitz stable attention module consisting of a dual orthogonal self-attention (DOSA) and a hierarchical channel cross-attention (HC2A) module. The main reason to ensure Lipschitz stability is to secure the robustness of the model against different adversarial attacks as discussed in \cite{4000, 4013}. We devise this whole pipeline under an adversarial setup as this enhances the overall outcomes of the model in terms of image segmentation by reducing the possibility of overfitting \cite{4014}. In the next section, we discuss a detailed architecture overview of our model as well as the required loss formulation for effective performance.

\begin{figure*}
    \centering
    \includegraphics[width=\textwidth]{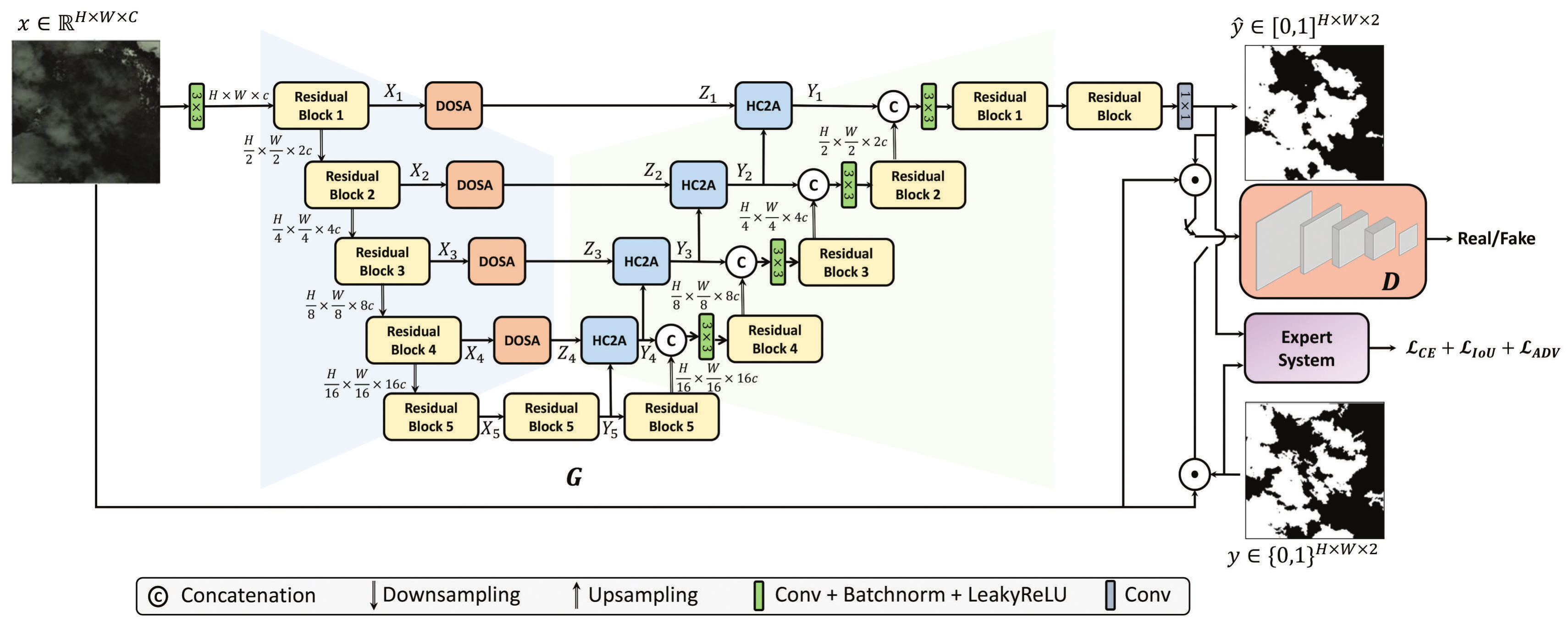}
    \caption{Overview of overall generative framework with proposed architecture of hierarchical hybrid transformer}
    \label{fig2}
\end{figure*}
\section{Methodology}\label{method}
\begin{figure*}
    \centering
    \includegraphics[width=\textwidth]{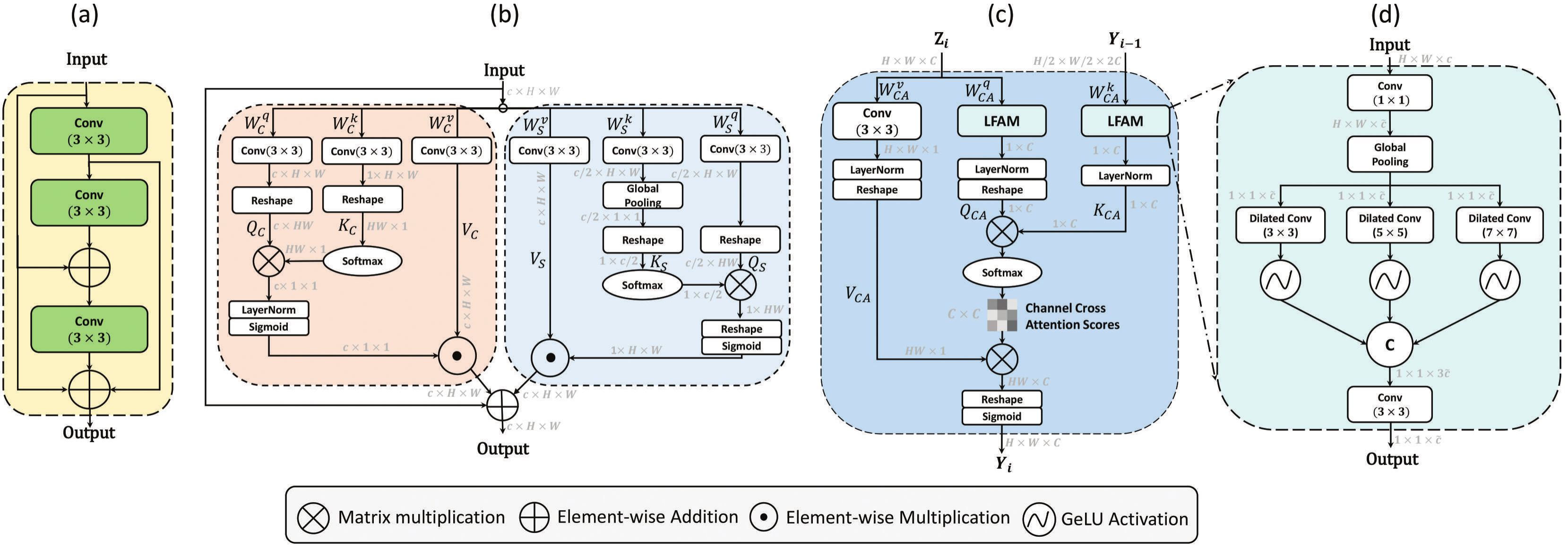}
    \caption{Overview of proposed attention modules and their components: (a) Residual block, (b) dual orthogonal self attention (DOSA), (c) hierarchical cross channel attention (HC2A), and (d) Laplacian feature aggregation module (LFAM).}
    \label{fig3}
\end{figure*}

Figure \ref{fig2} demonstrates the overview of our proposed framework followed by a detailed architectural outline of each module in Figure \ref{fig3}. The generator model $G$ operates on multispectral (MX) satellite image patches $\mathbf{x}\in \mathbb{R}^{H \times W \times C}$ where  $C$ corresponds to the number of channels in the input patch with height $H$ and width $W$. Let $\mathbf{x}\sim \mathbb{P}_{X}$,  where $\mathbb{P}_{X}$ is the joint source distribution depending on $C$. Let $\mathbf{y}\sim \mathbb{P}_{Y}$ be the target distribution with $\mathbf{y}\in \{0,1\}^{H\times W \times n}$, where $n$ is the total number of classes for segmentation and, let $\mathbf{\hat{y}}\sim \mathbb{P}_{G_\theta}$ where $\mathbf{\hat{y}} = G_{\theta}(x)$ with $\mathbf{\hat{y}} \in [0,1]^{H\times W \times n}$ and $ \mathbb{P}_{G_\theta}$ be the generator distribution parameterized by $\theta \in \Theta$. The discriminator $D$ is designed to classify ground truth data samples and generated data samples to be coming from real and fake sample space, respectively. In later sections, we will discuss a detailed architectural and optimization-oriented overview.
\subsection{Network Architecture}
In this study, we design a novel hierarchical hybrid transformer architecture based on the notion of the backbone structure of the U-Net \cite{1} model by incorporating with adversarial framework as shown in figure \ref{fig2}. For $D$, we borrowed the concept of PatchGAN \cite{3}, where model $G$ comprises an encoding part and a decoding part similar to the U-Net model. The main role of this encoding part is to extract high-resolution features and dense contextual information for input images while the decoding part works on these features to generate an output segmentation map. Unlike traditional U-Net, both of these components are incorporated with residual blocks along with upsampling and downsampling operations as shown in Figure \ref{fig2}. However, analogous to U-Net, we also use skip connections to assimilate the encoding and decoding feature maps to assist the decoder in producing precise and effective feature maps as well as to help in prompt gradient flow and context propagation in the network resulting in resolving vanishing gradient issues. Instead of straightforward skip connections, we augment it by introducing attention modules inspired by the multi-Dconv head transposed attention (MDTA) of Restormer \cite{21} to suppress activations in irrelevant regions, reducing the number of redundant features, especially for initial layers. We present two novel attention modules namely, Dual Orthogonal Self Attention (DOSA) to extract long-range structural and spectral information from the feature maps, and Hierarchical Cross Channel Attention (HC2A) to enhance the low-level residual features of skip connections by incorporating dense features from the deeper network. In the following section, we have provided a detailed overview of these two modules.
\subsubsection{Dual Orthogonal Self Attention (DOSA)}
The main purpose of DOSA is to capture long-range spectral and structural information in global context scenarios and update the weights of feature maps accordingly to share substantial information with every pixel. This results in an effective receptive field including the whole input image for which the decision for a certain pixel can be affected by spatial and spectral information of any other input pixel.  DOSA is designed to preserve high-resolution semantics during computing attention in pixel-wise regression compared to other self-attention modules available in the literature. For leveraging important features in both channel and spatial direction, it simultaneously computes the attentions in two orthogonal brunches, spatial-only-attention $(A^{sp}(\cdot))$ and channel-only attention $(A^{ch}(\cdot))$. Each brunch is inspired from \cite{29} and designed from the concept of polarizing filter in photography by completely collapsing feature maps in one direction while preserving high resolution in its orthogonal direction and also HDR (High Dynamic Range) operation by using Softmax and Sigmoid normalization for the smallest feature map to recover crucial details. The intrinsic architectural details of DOSA are demonstrated in Figure \ref{fig3}.\par
In both attention brunches, the query and key matrix are estimated as orthogonal to each other. In channel-only-attention brunch, for an input $\mathbf{X} \in \mathbb{R}^{H \times W \times c}$, the query and key are computed as $\mathbf{Q_{C}} \in \mathbb{R}^{HW \times c}$ and $\mathbf{K_{C}} \in \mathbb{R}^{1 \times HW}$, respectively. For input $\mathbf{X}$, the channel only attention is estimated such that $\mathbf{A^{ch}}(\mathbf{X}) \in \mathbb{R}^{1 \times 1 \times c}$ and, it is computed as:
\begin{equation}\label{101}
\begin{aligned}
\mathbf{A}^{c h} (\mathbf{X})= \sigma (F_{RS} [\mathbf{W}^{q}_C * \mathbf{X}] \times \phi(F_{RS} [\mathbf{W}^{k}_C * \mathbf{X}])),
\end{aligned}
\end{equation}
where $\mathbf{W}^{q}_C$ and $\mathbf{W}^{K}_C$ are $3\times 3$ convolution operations with the corresponding weight matrix of query and key with $*$ being the convolution operation, $F_{RS}$ is reshape operator, and $\sigma(\cdot)$ and $\phi(\cdot)$ are Sigmoid and Softmax functions, respectively. The output of the channel-only-attention is estimated as \par
Similarly, for the spatial-only-attention, the corresponding query $(\mathbf{Q_{S}})$ and key $(\mathbf{K_{S}})$ are estimated such that $\mathbf{Q_{S}} \in \mathbb{R}^{HW \times c}$ and $\mathbf{K_{S}} \in \mathbb{R}^{1 \times C}$, and the related attention $\mathbf{A^{sp}}(\mathbf{X}) \in \mathbb{R}^{H \times W \times 1}$ is approximated as:
\begin{equation}\label{eq102}
\begin{aligned}
\mathbf{A}^{s p}(\mathbf{X})=\sigma(F_{RS}\bigl[\phi(F_{RS}\left[F_{G P}  (\mathbf{W}^{k}_S* \mathbf{X})\right]) \\ 
 \times F_{RS} \left[\mathbf{W}^{q}_S * \mathbf{X} \right]\bigr]),    
\end{aligned}
\end{equation}
where $F_{GP}(\cdot)$ is a global pooling operator $F_{G P}(\mathbf{X}) = \frac{1}{H \times W} \sum_{i=1}^H \sum_{i=1}^W X(:, i, j)$, $\mathbf{W}^{q}_S$, $\mathbf{W}^{k}_S$ are $3\times 3$ convolution operations with the corresponding weight matrix of query, and key during estimating spatial attention. 
The end product of DOSA can be conveyed with a parallel layout of the outputs of the above two branches as shown in equation \ref{eq4}.
\begin{equation}\label{eq4}
\begin{aligned}
\mathbf{Z} = \mathbf{X}+\mathbf{A}^{c h}(\mathbf{X}) \odot (\mathbf{W^{v}_C} *\mathbf{X})+\mathbf{A}^{s p}(\mathbf{X}) \odot (\mathbf{W^{v}_S} * \mathbf{X}),
\end{aligned}
\end{equation}
where $\mathbf{Z}=DOSA(\mathbf{X})$, $\mathbf{W^{v}_C}$ and $\mathbf{W^{v}_S}$ are $3\times 3$ convolution kernels corresponding to value matrix $\mathbf{V_C}$ and $\mathbf{V_S}$ for channel and spatial attention branch, respectively and $\odot$ denotes element-wise multiplication. \par
Apart from extracting long range dependencies, the motivation behind DOSA is to provide better sensitivity in terms of Lipschitz constant. As shown in \cite{4007}, the $L_2$-Lipschitz constant of self attention mechanism in traditional transformer model is bounded by the variance of the input, resulting in larger sensitivity. In this context, we propose \textbf{Theorem 1} to derive the approximate Lipschitz bound of the DOSA module to visualize its sensitivity with respect to any given input. \par
\begin{thm}
    \textit{Let $m$ and $M$ be the minimum and maximum of input feature $\mathbf{X}\in \mathbb{R}^{H\times W \times c}$. Let $\omega=e^{\frac{2\pi i}{HW}}$ and $W^q$ be the convolution kernel for query matrix in DOSA. Let $\Gamma$ be the difference between the learned kernel and its initialization and given by $\Gamma = W^q - W^q_{0}$. Let $U$ be a complex matrix such that $U_{ij}=\omega^{ij}$. If $\varepsilon^{q} = \frac{1}{9}(U^T \Gamma U)_{0,0}$, then the upper bound on the magnitude of diagonal elements of Jacobian of DOSA is estimated as,}
\begin{equation}\label{eq40}
\begin{aligned}
||J^{SA}_{i,i}||_2 \leq & 1+\frac{1}{2}\bigl(1+ 9\varepsilon_i^{v}\bigr)\bigl(1+ 9\varepsilon_i^{q}\bigr)\bigl(B \\
& + \sqrt{2}(1+9\varepsilon_i^{k})B^3\bigr) + 2\bigl(1+ 9\varepsilon_i^{v}\bigr),
\end{aligned}
\end{equation} 
\textit{where, $B = \sqrt{HW}\max(||m||,||M||)$, $\varepsilon_i^{v}$ and $\varepsilon_i^{k}$} are corresponding to value and key matrix in DOSA.
\end{thm}
\begin{proof}
This proof is inspired from \cite{4007} and an immediate application of operator norm for convolution kernels \cite{4000} which says 2-D convolution operations can be represented as matrix multiplication with corresponding operator matrix, i.e. $W_f*x \overset{\mathrm{def}}{=} op(W_f)x$. Therefore, the output of the channel attention branch using equation \ref{101} can be rewritten as,
\begin{equation}\label{eq41}
\begin{split}
\mathbf{Z^{ch}} &= \mathbf{A}^{c h}(\mathbf{X}) \odot (op(\mathbf{W^{v}_C})\mathbf{X}) \\
&= \begin{aligned}[t]
    \sigma (F_{RS} [op(\mathbf{W}^{q}_C)\mathbf{X}] & \times \phi(F_{RS} [op(\mathbf{W}^{k}_C)\mathbf{X}])) \\
& \odot (op(\mathbf{W^{v}_C})\mathbf{X})
\end{aligned}
\end{split}
\end{equation}
To estimate the magnitude of Jacobian $\mathbf{Z^{ch}}$, the reshape operator $F_{RS}(\cdot)$ can be ignored, and hence, we can write,
\begin{equation}\label{eq41}
\begin{aligned}
||J^{ch}_{i,i}|| & =  ||\frac{\partial \mathbf{Z_i^{ch}}}{\partial \mathbf{X}_i}||=||\Bigl(diag(\sigma(\mathbf{\alpha_1})) \odot op(\mathbf{W^{v}_{C_i}}) \Bigr)\\
& + \Bigl(diag(op(\mathbf{W^{v}_{C_i}})\mathbf{X}_i \odot \sigma(\mathbf{\alpha_1})(1-\sigma(\mathbf{\alpha_1}))) \Bigr)\\
&\times \Bigr(diag(op(\mathbf{W}^{q}_{C_i}).\phi(\beta_1) +  op(\mathbf{W}^{q}_{C_i})\mathbf{X}_i. \frac{\partial}{\partial \mathbf{X}_i} \phi(\beta_1))\Bigl)||,
\end{aligned}
\end{equation} 
where, $\alpha_1 = op(\mathbf{W}^{q}_C)\mathbf{X} \times \phi(op(\mathbf{W}^{k}_C)\mathbf{X})$ and $\beta_1 = op(\mathbf{W}^{k}_C)\mathbf{X}$. Now using triangular and Cauchy-Schwarz inequality to equation \ref{eq41}, we will get,
\begin{equation}\label{eq42}
\begin{aligned}
&||J^{ch}_{i,i}|| \leq ||\Bigl(diag(\sigma(\mathbf{\alpha_1})) \odot op(\mathbf{W^{v}_{C_i}}) \Bigr)||\\
 & + ||\Bigl(diag(op(\mathbf{W^{v}_{C_i}})\mathbf{X}_i \odot \sigma(\mathbf{\alpha_1})(1-\sigma(\mathbf{\alpha_1}))) \Bigr)||\\
& \times \Bigr(||diag(op(\mathbf{W}^{q}_{C_i}).\phi(\beta_1))|| +  ||op(\mathbf{W}^{q}_{C_i})\mathbf{X}_i. \frac{\partial}{\partial \mathbf{X}_i} \phi(\beta_1)||\Bigl),
\end{aligned}
\end{equation} 
From \cite{4001}, we know for any two  matrix $\mathbf{A}\in \mathbb{R}^{d_1\times n}$ and $\mathbf{B}\in \mathbb{R}^{d_2\times n}$ dependent on X, the following inequality holds,  
\begin{equation}\label{eq43}
\begin{aligned}
||\mathbf{A} \frac{\partial}{\partial \mathbf{X}} \text{softmax}(\mathbf{B})||_2 \leq \sqrt{2} ||\mathbf{A}||_{(2,\infty)}||\mathbf{B}||_{(\infty,2)},
\end{aligned}
\end{equation} 
where, $||\mathbf{B}||_{(\infty,2)} = \max_i (\sum_{j}B^2_{i,j})^{1/2}$ \cite{4001}. Then considering the norm in equation \ref{eq42} as L2, and taking the operator norm for a $t \times t \times 1$ shaped kernel $W_f$ as $||op(W_f)||_2 = 1+ \varepsilon t^2$, the euclidean Lipschitz constant from equation \ref{eq42} can be estimated as,
\begin{equation}\label{eq44}
\begin{aligned}
||J^{ch}_{i,i}||_2 & \leq \frac{1}{4}\bigl(1+ 9\varepsilon_i^{v^C}\bigr)\bigl(1+ 9\varepsilon_i^{q^C}\bigr)\bigl(||X_i||_2 \\
& + \sqrt{2}(1+9\varepsilon_i^{k^C})||X_i||_2^3\bigr) + \bigl(1+ 9\varepsilon_i^{v^C}\bigr).
\end{aligned}
\end{equation} 
Similarly for the spatial attention branch, the corresponding eucledian norm of the Jacobian from equation \ref{eq102} can be estimated as, 
\begin{equation}\label{eq45}
\begin{aligned}
||J^{sp}_{i,i}||_2 & \leq \frac{1}{4}\bigl(1+ 9\varepsilon_i^{v^S}\bigr)\bigl(1+ 9\varepsilon_i^{q^S}\bigr)\bigl(||X_i||_2 \\
& + \sqrt{2}(1+9\varepsilon_i^{k^S})||X_i||_2^3\bigr) + \bigl(1+ 9\varepsilon_i^{v^S}\bigr).
\end{aligned}
\end{equation}
For simplicity, we assume the convolutional kernels regarding key, query and value matrix are identical for both channel and spatial attention. Then taking $||X||_2 \leq \sqrt{HW}\max(||m||,||M||)$, the final euclidean Lipschitz constant of DOSA can be approximated using equation \ref{eq4},

\begin{equation}\label{eq46}
\begin{aligned}
||J^{SA}_{i,i}||_2  = ||\frac{\partial \mathbf{Z_i}}{\partial \mathbf{X}_i}||_2 & \leq 1+\frac{1}{2}\bigl(1+ 9\varepsilon_i^{v}\bigr)\bigl(1+ 9\varepsilon_i^{q}\bigr)\bigl(B \\
& + \sqrt{2}(1+9\varepsilon_i^{k})B^3\bigr) + 2\bigl(1+ 9\varepsilon_i^{v}\bigr),
\end{aligned}
\end{equation}
which completes the proof.
\end{proof}
\textbf{Theorem 1} shows unlike traditional self attention modules, the euclidean Lipschitz constant of DOSA is upper bounded by the cubical of the absolute maximum of input. Therefore, its performance is relatively stable for $[-1,1]$ normalized input. From the perspective of our designed model, this is crucial as almost all input features to DOSA are $[-1,1]$ bounded because of batcnorm and layernorm utilization. The output of DOSA is passed through HC2A block, which we discuss in next section.
\subsubsection{Hierarchical Cross Channel Attention (HC2A)}
Although DOSA ensures to intertwines the skip connection activation maps with respect to extracted long range structural and spectral information, it does not guarantee to enhance the poor feature representations in the initial layers in terms of semantic richness. However, the feature maps in the deeper network comprise of dense semantic information that can be utilized to update the initial feature maps resulting in escalation of efficacy of the decoder in Figure \ref{fig2}. HC2A module is designed based on this perception and its key aspect is to actively suppress activations in irrelevant regions, reducing the number of redundant features brought across and highlight areas that present significant interest to the application. HC2A module acts in a hierarchical manner, in which at particular downsampling level $n$, it generated the enhanced feature activation maps as $\mathbf{Y_n} = HC2A(\mathbf{Z_n}, \mathbf{Y_{n-1}})$ as shown in Figure \ref{fig2}. The detailed architectural details of HC2A module is illustrated in Figure \ref{fig3}(c).\par
The HC2A block consists of a Laplacian Feature Aggregation Module (LFAM), which is built upon a Laplacian pyramid structure with $3\times 3$ convolution operations with dilation rate 3, 5, 7 followed by GeLU activation functions. We concatenate these feature maps and apply a $3\times 3$ convolution operation as visualized in Figure \ref{fig3}(d). By employing LFAM, we not only ensure generation of sparse channel activation maps by eliminating noisy spectral information in latent space, but also assure efficient computation by being independent of any spatial feature manipulation. This is important as deeper feature maps contain their dense semantic information in their spectral direction. \par
As shown in Figure \ref{fig3}(c), at particular downsampling level $n$, the query matrix $\mathbf{Q_{CA}}$ is computed by passing input $\mathbf{Z_n} \in \mathbb{R}^{H\times W \times C}$ through a LFAM block such that $\mathbf{Q_{CA}} \in \mathbb{R}^{1 \times C}$. Similarly, the key matrix $\mathbf{K_{CA}}$ is estimated for input $\mathbf{Y_{n-1}} \in \mathbb{R}^{H/2\times W/2 \times 2C}$ such that $\mathbf{K_{CA}} \in \mathbb{R}^{1 \times C}$. Therefore, the output of the cross attention block is formulated as in equation \ref{eq5}.
\begin{equation}\label{eq5}
\begin{split}
    \mathbf{Y_n} = \sigma \bigl(F_{RS} \bigl[ F_{RS}\left[\mathbf{W^v_{CA}} * \mathbf{Z_n}\right] & \times \phi \bigl( LFAM(\mathbf{Z_n}) \\ 
    &  \times LFAM(\mathbf{Y_{n-1}})\bigr)\bigr]\bigr)
\end{split}
\end{equation}
where, $\mathbf{W^v_{CA}}$ is $3\times 3$ is the convolution kernel to estimate the value matrix $\mathbf{V_{CA}}$ from input $\mathbf{Z_n}$. 
Similar to \textbf{Theorem 1}, we can approximate the Lipschitz constant of HC2A and based on that, we can investigate how it affects the sensitivity of overall attention module in the skip connection. In this context we propose \textbf{Theorem 2}.
\begin{thm}
    \textit{Using the definition of operator norm in \textbf{Theorem 1}, the bound  on the magnitude of diagonal elements of Jacobian of HC2A can be given by,}
    \begin{equation}
    \begin{aligned}
        ||J^{CA}_{i,i}|| \leq &   \frac{1}{2\sqrt{2}}(1+9\varepsilon_{i,n}^{v^{CA}})\bigl(\frac{1}{\sqrt{2}}+(1+9\varepsilon_{i,n}^{q^{CA}})\\
        & (1+9\varepsilon_{i,n}^{k^{CA}})||\mathbf{Z_{i,n}}||^2.||\mathbf{Y_{i,n-1}}||\bigr),
    \end{aligned}
\end{equation}
\textit{where, $\varepsilon_{n}^{q^{CA}}$, $\varepsilon_{n}^{k^{CA}}$ and $\varepsilon_{n}^{v^{CA}}$ corresponds to query, key and value matrix in HC2A. Then, the overall Jacobian of DOSA $+$ HC2A is bounded $\mathcal{O}(B^3 + B^4 + B^6)$ with $B = \sqrt{HW}\max(||m||,||M||)$ for an input $\mathbf{X_n}\in \mathbb{R}^{H\times W \times C}$.}
\end{thm}
\begin{proof}
    For simplicity, we assume the operation of LFAM similar to $3\times 3$ convolution operations with $\mathbf{W}^q_{CA}$ and $\mathbf{W}^k_{CA}$ kernels for key and query matrix as shown in Figure \ref{fig3} (a). Then using the similar operator norm definition of convolution kernel, the cross attention can be rewritten as,
\begin{equation}\label{eq6}
\begin{split}
    \mathbf{Y_n} = \sigma \bigl(F_{RS} \bigl[ F_{RS}\left[op(\mathbf{W^v_{CA}}) \mathbf{Z_n}\right] &\times \phi \bigl( op(\mathbf{W}^q_{CA})\mathbf{Z_n} \\ 
    &  \times op(\mathbf{W}^k_{CA})\mathbf{Y_{n-1}}\bigr)\bigr]\bigr)
\end{split}
\end{equation}\label{eq7}
Then ignoring the reshape operator, the magnitude of the diagonal elements of Jacobian with respect to input $\mathbf{Z_n}$ can be written as,
\begin{equation}
    \begin{aligned}
        ||J^{CA}_{i,i}|| = &||\frac{\partial \mathbf{Y_{i,n}}}{\partial \mathbf{Z_{i,n}}}|| = ||\sigma(\alpha_2)(1-\sigma(\alpha_2)) \\
        &\times \Bigl[op(\mathbf{W^v_{CA,i}}) \times diag(\phi(\beta_2))\\
        &+ op(\mathbf{W^v_{CA,i}}) \times diag\bigl(\mathbf{Z_{i,n}}\frac{\partial}{\partial \mathbf{Z_{i,n}}}\phi(\beta_2)\bigr)\Bigr]||
    \end{aligned}
\end{equation}
where, $\alpha_2 = \left[op(\mathbf{W^v_{CA,i}}) \mathbf{Z_{i,n}}\right]\times \phi(\beta_2)$  and $\beta_2=op(\mathbf{W^q_{CA,i}})\mathbf{Z_{i,n}}\times op(\mathbf{W^k_{CA,i}})\mathbf{Y_{i,n-1}}$. Now using the definition of operator norm from \textbf{Theorem 1} and using equation \ref{eq43}, we can write using triangular inequality,
\begin{equation}\label{eq8}
    \begin{aligned}
        ||J^{CA}_{i,i}|| \leq &   \frac{1}{2\sqrt{2}}(1+9\varepsilon_{i,n}^{v^{CA}})\bigl(\frac{1}{\sqrt{2}}+(1+9\varepsilon_{i,n}^{q^{CA}})\\
        & (1+9\varepsilon_{i,n}^{k^{CA}})||\mathbf{Z_{i,n}}||^2.||\mathbf{Y_{i,n-1}}||\bigr),
    \end{aligned}
\end{equation}
As almost all the feature maps are bounded in $[-1,1]$ due to usage of layer-norms, and $\mathbf{Z_n} = DOSA(\mathbf{X_n})$ from figure \ref{fig2}, we can assume ||$\mathbf{Z_n}||_2\leq B$ and $||\mathbf{Y_{n-1}}||_2\leq \frac{B}{2}$. Therefore, we can simply rewrite equation \ref{eq8} as $||J^{CA}_{i,i}||= ||\frac{\partial \mathbf{Y_{i}}}{\partial \mathbf{Z_{i}}}|| \leq \mathcal{O}(B^3)$. Similarly, we can simplify equation \ref{eq46} to write $||J^{SA}_{i,i}||=||\frac{\partial \mathbf{Z_{i}}}{\partial \mathbf{X_{i}}}|| \leq \mathcal{O}(1+B+B^3)$. Then, to approximate the magnitudes on diagonals of Jacobian for the overall DOSA+HC2A module, we can formulate it using chain rule, and Cauchy-Schwarz inequality as shown in equation \ref{eq9}.
\begin{equation}\label{eq9}
    \begin{aligned}
        ||J_{i,i}||_2 &= ||\frac{\partial \mathbf{Y_{i,n}}}{\partial \mathbf{X_{i,n}}}||_2 \leq ||\frac{\partial \mathbf{Y_{i,n}}}{\partial \mathbf{Z_{i,n}}}||_2 ||\frac{\partial \mathbf{Z_{i,n}}}{\partial \mathbf{X_{i,n}}}||_2 \\
        & \leq \mathcal{O}(B^3).\mathcal{O}(1+B+B^3)\\
        &= \mathcal{O}(B^3 + B^4 + B^6),
    \end{aligned}
\end{equation}
which completes the proof.
\end{proof}
\textbf{Theorem 2} shows that besides providing dense features to accurately generate segmentation map, it improves the overall Lipschitz sensitivity which makes it more substantially stable for normalized $[-1,1]$ bounded input. Later, we have provided considerable evidence, how this affects the overall performance of the model.

\subsection{Network objective definition}\label{obj}
The overall framework is optimized by using a set of loss functions along with the minmax objective function of conditional GAN \cite{3}. In the following sections, we define the objective functions to train generator $G$ and discriminator $D$.
\subsubsection{Discriminator Objective Function}
Let the discriminator model $D$ be parameterized by $\psi \in \Psi$. As discussed above, $D_\psi$ is trained based on the concept of conditional GAN \cite{3}. Therefore, $D_\psi$ will take generated mask $\mathbf{\hat{y}}$ and true label $\mathbf{y}$, both conditioned on input $\mathbf{x}$ and classify the corresponding patch as real and fake respectively. Thus, the objective function of $D_\psi$ can be defined as in equation (\ref{eq10}).
\begin{equation} \label{eq10}
\begin{aligned}
\min _\mathbf{\psi} - \mathbb{E}_{\mathbf{x}\sim \mathbb{P}_X, \mathbf{y} \sim \mathbb{P}_Y}&[\log (D_{\psi}(\mathbf{x},\mathbf{y}))]\\ - 
& \mathbb{E}_{\mathbf{x}\sim \mathbb{P}_X, \mathbf{\hat{y}} \sim \mathbb{P}_{\mathbf{G}_\theta}}[\log (1-D_{\psi}(\mathbf{x}, \mathbf{\hat{y}}))].     
\end{aligned}
\end{equation}

\subsubsection{Generator Objective Function}
The generator model $G$, parameterized with $\theta \in \Theta$, yields segmentation map $\mathbf{\hat{y}}$. The prime objective of $G$ is to generate $\mathbf{\hat{y}}$ closely related to the true label $\mathbf{y}$ while maintaining underlying statistical attributes close to the real manifold. For this purpose, $G_\theta$ is trained with a combination of Focal loss and Lov\'asz-Softmax loss to generate a faithful segmentation map, while to coincide statistical measures between $y$ and $\hat{y}$, it is also regularized with adversarial loss. Below we discuss the role and implication of each loss to provide optimum solution.\par

\textbf{Focal loss:} The main role of Focal loss is similar to cross-entropy (CE) loss which provides a measure of dissimilarities between predicted probabilities and ground truth labels. The major advantage of focal loss over CE loss is its ability to focus on examples that are difficult to classify like thin clouds while reducing the influence of easily classified ones. This is done by tuning a focusing parameter $\gamma$ in the following loss function.  
\begin{equation} \label{eq11}
\begin{aligned}
\mathscr{L}_{CE} = \mathbb{E}_{\mathbf{y} \sim \mathbb{P}_Y, \mathbf{\hat{y}}\sim \mathbb{P}_{\mathbf{G}_\theta}} -w(1-\mathbf{\hat{y}})^\gamma\log (\mathbf{\hat{y}}),
\end{aligned}
\end{equation}
where, $w\in \mathbb{R}^n$ corresponds to weights for $n$ number of classes to handle class imbalance during loss computation. Focal loss similar to CE 
 is very sensitive to outliers or mislabeled classes resulting in
poor generalization ability. Except this, CE loss is unable to accurately delineate object boundary.
To overcome these limitations, we have incorporated this loss with Lov\'asz-Softmax loss to train $G$.\par

\textbf{Lov\'asz-Softmax Loss:}
Jaccard index, also referred as intersection-over-union (IoU) score, is one of the key performance measure in semantic segmentation suggesting the amount of intersection between set of true labels and generated labels. Therefore, Jaccard index is often used as loss function for not only boundary enhancement but also precise location of the generated segmentation map. For ground truth label $\mathbf{y}$ and generated label $\mathbf{\hat{y}}$, the Jaccard loss $(\Delta_{\mathcal{J}_c})$ for a class $\mathrm{c}\in \mathcal{C}$ can be written as a set of mispredictions $(\mathbf{M}_c)$ as shown below.
\begin{equation} \label{eq14}
\begin{aligned}
\mathbf{M}_c(\mathbf{y}, \mathbf{\hat{y}}) = \{\mathbf{y} =\mathrm{c}, \mathbf{\hat{y}}\neq \mathrm{c}\}\cap \{\mathbf{y} \neq \mathrm{c}, \mathbf{\hat{y}}= \mathrm{c}\}\\
\Delta_{\mathcal{J}_c}: \mathbf{M}_c \in \{0,1\}^p \mapsto \frac{|\mathbf{M}_c|}{|\{\mathbf{y} = \mathrm{c}\} \cap \mathbf{M}_c|},
\end{aligned}
\end{equation}
where $p$ is the number of pixels in the minibatch. However, as $(\Delta_{\mathcal{J}_c})$ is a set function, its optimization is not easily interactable due to its non-smooth properties. Therefore, for effective convergence of the model, we apply convex surrogate of Jacard loss by implying Lov\'asz extension to ensure smoothness and differentiability as shown in \cite{4003} which is defined in equation \ref{eq15}.
\begin{equation} \label{eq15}
\begin{aligned}
\mathscr{L}_{\mathcal{J}} = \mathbb{E}_{\mathbf{y} \sim \mathbb{P}_Y, \mathbf{\hat{y}}\sim \mathbb{P}_{\mathbf{G}_\theta}} \frac{1}{|\mathcal{C}|}\sum_{\mathrm{c}\in \mathcal{C}} \overline{\Delta_{\mathcal{J}_c}} (\phi(\mathbf{M}_c(\mathbf{y}, \mathbf{\hat{y}}))), \\
\text{where} \medspace \overline{\Delta_{\mathcal{J}_c}} : \textbf{m} \in \mathbb{R}^p \mapsto \sum_{i=1}^{p} m_i g_i(\mathbf{m}), \: \mathbf{m} = \phi(\mathbf{M}_c(\cdot)),\\
\text{with} \medspace g_i(\mathbf{m}) = \Delta_{\mathcal{J}_c}(\{\pi_1,\ldots, \pi_i\}) - \Delta_{\mathcal{J}_c}(\{\pi_1,\ldots, \pi_{i-1}\}),
\end{aligned}
\end{equation}
where, $\phi$ is softmax operation, and $\mathbf{\pi}$ being a permutation ordering the oder of $\mathbf{m}$ in decreasing order such that $x_{\pi_1} \geq x_{\pi_2} \ldots \geq x_{\pi_p}.$  

\begin{table*}[]
\caption{Quantitative analysis on Biome dataset and comparison with other state-of-the-art methods. First and second methods are highlighted in red and green, respectively.}
\label{tab:my-table}
\resizebox{\textwidth}{!}{%
\begin{tabular}{c|ccccccc|ccccccc}
\hline
Band Configuration & \multicolumn{7}{c|}{C=1}                                                                                                                                                                                                 & \multicolumn{7}{c}{C=3}                                                                                                                                                                                                 \\ \hline
Method             & OA                           & Rec                          & Pre                          & mIoU(\%)                     & Kappa                        & BI(\%)                       & HD (m)                         & \multicolumn{1}{c}{OA}      & Rec                          & Pre                          & mIoU(\%)                     & Kappa                        & BI(\%)                       & HD (m)                        \\ \hline
CDNetV2 \cite{20}           & 92.95                        & 90.31                        & 91.95                        & 89.03                        & 73.76                        & 36.35                        & 1911.76                        & 91.28                        & 87.67                        & 90.38                        & 86.86                        & 77.29                        & 32.49                        & 1926.86                       \\
CDNetV1 \cite{19}           & 88.79                        & 88.31                        & 86.78                        & 84.83                        & 73.65                        & 36.81                        & 1980.17                        & 94.46                        & {\color[HTML]{009901} \textit{94.97}} & 92.17                        & 91.64                        & 86.77                        & 40.19                        & 1508.82                       \\
DeepLabV3+ \cite{23}        & 91.54                        & 90.58                        & 90.13                        & 88.29                        & 80.12                        & 39.89                        & 1683.65                        & 91.82                        & 91.43                        & 90.21                        & 88.61                        & 81.17                        & 39.92                        & 1511.43                       \\
PSPnet \cite{22}            & 92.99                        & 92.55                        & 90.67                        & 89.68                        & 83.08                        & {\color[HTML]{009901} \textit{39.70}} & 1727.19                        & 94.47                        & 94.19                        & 92.35                        & 91.06                        & 86.16                        & 39.36                        & 1556.73                       \\
RS-Net \cite{18}            & 90.21                        & 83.66                        & {\color[HTML]{009901} \textit{94.70}} & 87.28                        & 76.99                        & 33.01                        & 1723.82                        & 93.53                        & 92.42                        & 94.46                        & 91.09                        & 84.36                        & 36.16                        & 1508.07                       \\
RefUNetV4 \cite{4010}           & {\color[HTML]{009901} \textit{94.44}} & {\color[HTML]{009901} \textit{94.46}} & 92.02                        & {\color[HTML]{009901} \textit{91.59}} & {\color[HTML]{009901} \textit{85.84}} & 29.24                        & {\color[HTML]{009901} \textit{1365.91}} & {\color[HTML]{009901} \textit{95.83}} & {\color[HTML]{9A0000} \textbf{95.09}} & {\color[HTML]{009901} \textit{94.33}} & {\color[HTML]{009901} \textit{93.40}} & {\color[HTML]{009901} \textit{89.63}} & {\color[HTML]{009901} \textit{46.64}} & {\color[HTML]{009901} \textit{865.48}} \\
CLiSA (ours)               & {\color[HTML]{9A0000} \textbf{97.02}} & {\color[HTML]{9A0000} \textbf{96.31}} & {\color[HTML]{9A0000} \textbf{95.80}} & {\color[HTML]{9A0000} \textbf{94.77}} & {\color[HTML]{9A0000} \textbf{91.88}} & {\color[HTML]{9A0000} \textbf{55.19}} & {\color[HTML]{9A0000} \textbf{601.19}}  & {\color[HTML]{9A0000} \textbf{97.52}} & 89.33                        & {\color[HTML]{9A0000} \textbf{96.43}} & {\color[HTML]{9A0000} \textbf{95.29}} & {\color[HTML]{9A0000} \textbf{92.82}} & {\color[HTML]{9A0000} \textbf{56.90}} & {\color[HTML]{9A0000} \textbf{577.42}} \\ \hline
Band Configuration & \multicolumn{7}{c|}{C=4}                                                                                                                                                                                                 & \multicolumn{7}{c}{C=11}                                                                                                                                                                                                \\ \hline
Method             & OA                           & Rec                          & Pre                          & mIoU(\%)                     & Kappa                        & BI(\%)                       & HD (m)                         & \multicolumn{1}{c}{OA}      & Rec                          & Pre                          & mIoU(\%)                     & Kappa                        & BI(\%)                       & HD (m)                        \\ \hline
CDNetV2 \cite{20}           & 89.52                        & 91.56                        & 86.12                        & 84.56                        & 75.92                        & 34.68                        & 1961.52                        & 87.12                        & 77.15                        & 92.35                        & 82.01                        & 62.32                        & 31.20                        & 2637.38                       \\
CDNetV1 \cite{19}           & 94.38                        & 92.65                        & 93.62                        & 91.64                        & 86.62                        & 40.99                        & 1538.30                        & 94.34                        & 92.68 & 93.71                        & 91.47                        & 86.04                        & 40.47                        & 1568.82                       \\
DeepLabV3+ \cite{23}        & 93.48                        & 93.23                        & 92.07                        & 90.57                        & 85.03                        & 39.29                        & 1465.99                        & 93.47                        & 92.14                        & 92.50                        & 90.42                        & 84.75                        & 35.26                        & 1583.98                       \\
PSPnet \cite{22}            & 94.33                        & 94.74                        & 92.13                        & 91.39                        & 87.27                        & 39.4  & 1381.27                        & 95.11                        & {\color[HTML]{009901} \textit{94.08}} & 93.93                        & 92.30                        & 88.2                         & 41.21                        & 1344.44                       \\
RS-Net \cite{18}            & 92.11                        & 90.58                        &  91.66 & 89.30                        & 84.05                        & 42.58                        & 1321.87                        & 93.81                        & 91.58                        & {\color[HTML]{009901} \textit{95.10}} & 91.24                        & {\color[HTML]{009901} \textit{88.20}} & 44.91                        & 1178.96                       \\
RefUNetV4 \cite{4010}            & {\color[HTML]{009901} \textit{95.36}} & {\color[HTML]{009901} \textit{93.48}} & {\color[HTML]{009901} \textit{94.67}} & {\color[HTML]{009901} \textit{93.07}} & {\color[HTML]{009901} \textit{88.04}} & {\color[HTML]{009901} \textit{47.81}} & {\color[HTML]{009901} \textit{890.69}}  & {\color[HTML]{009901} \textit{95.33}} & 94.00 & 93.99 & {\color[HTML]{009901} \textit{93.98}} &  87.52 & {\color[HTML]{009901} \textit{47.03}} & {\color[HTML]{009901} \textit{941.49}} \\
CLiSA (ours)               & {\color[HTML]{9A0000} \textbf{97.43}} & {\color[HTML]{9A0000} \textbf{97.26}} & {\color[HTML]{9A0000} \textbf{95.85}} & {\color[HTML]{9A0000} \textbf{95.17}} & {\color[HTML]{9A0000} \textbf{92.81}} & {\color[HTML]{9A0000} \textbf{57.71}} & {\color[HTML]{9A0000} \textbf{553.15}}  & {\color[HTML]{9A0000} \textbf{98.05}} & {\color[HTML]{9A0000} \textbf{97.78}} & {\color[HTML]{9A0000} \textbf{96.88}} & {\color[HTML]{9A0000} \textbf{96.02}} & {\color[HTML]{9A0000} \textbf{94.09}} & {\color[HTML]{9A0000} \textbf{59.33}} & {\color[HTML]{9A0000} \textbf{497.12}} \\ \hline
\end{tabular}%
}
\end{table*}
\begin{figure*}[ht]
\centering
\includegraphics[width=\textwidth, height=6.5cm]{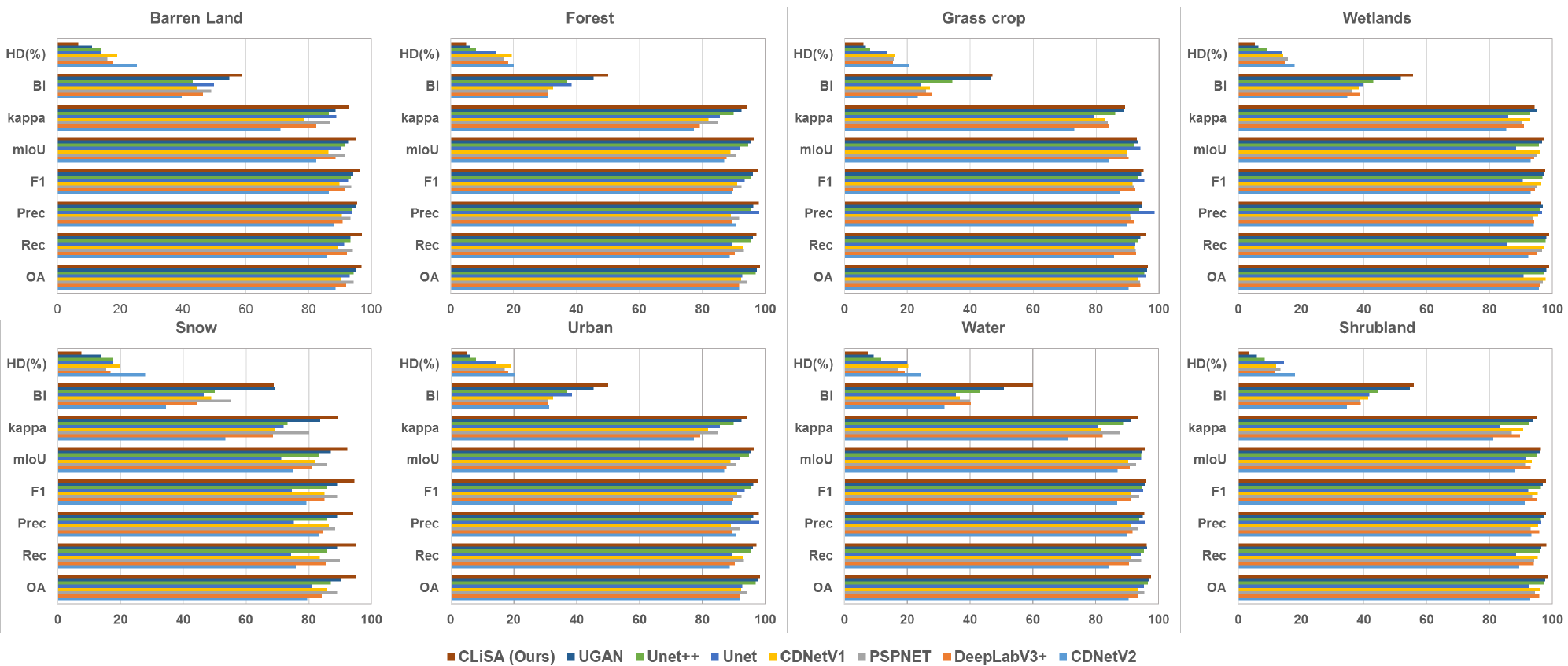}
\caption{Comparative quantitative analysis on cloud mask generation in presence of various signatures.}
\label{fig:signature}
\end{figure*}

\textbf{Adversarial Loss:} The primary objective of this loss is to help $G_\theta$ to model the underlying distribution of the ground truth labels. This will ensure the generated segmentation mask and true mask labels are statistically identical which will help $G$ to generate mask more accurately and close to the real manifold. The adversarial loss
is defined as in equation \ref{eq12}.
\begin{equation}\label{eq12}
\mathscr{L}_{ADV} = \mathbb{E}_{\mathbf{x}\sim \mathbb{P}_X, \mathbf{\hat{y}} \sim \mathbb{P}_{\mathbf{G}_\theta}} -\log (D_\psi(G_\theta(\mathbf{x}, \mathbf{\hat{y}})).
\end{equation}\par
Along with the above mentioned losses, we have comprised the generator objective function by regularizing it with L2 regression with respect to the parameters of $G$ to avoid overfitting. Therefore, the final objective for training the generator can be defined as in equation \ref{eq13}.
\begin{equation}\label{eq13}
\min _\mathbf{\theta} \lambda_{CE}\mathscr{L}_{CE} + \lambda_{IoU}\mathscr{L}_{\mathcal{J}} + \lambda_{ADV}\mathscr{L}_{ADV} + \lambda_{L2}.||\theta||_{2}^{2},
\end{equation}
where, $\lambda_{CE}$, $\lambda_{IoU}$, $\lambda_{ADV}$, and $\lambda_{L2}$ represent the weights assigned to Focal loss, Lov\'asz-Softmax Loss, adversarial loss and tikonov regularization, respectively.
\begin{figure*}[]
\centering
\begin{subfigure}{0.009\linewidth}
    \caption*{a.}
    \vspace{1.4cm}
    \caption*{b.}
    \vspace{1.4cm}
    \caption*{c.}
    \vspace{1.4cm}
    \caption*{d.}
    \vspace{1.24cm}
    \caption*{e.}
    \vspace{1.28cm}
    \caption*{f.}
    \vspace{1.3cm}
    \caption*{g.}
    \vspace{1.38cm}
    \caption*{h.}
    \vspace{1.4cm}
\end{subfigure}
\begin{subfigure}{0.1\linewidth}
    \caption*{Input Image} 
    \adjustbox{width=\linewidth, frame=0pt 1pt}{\includegraphics{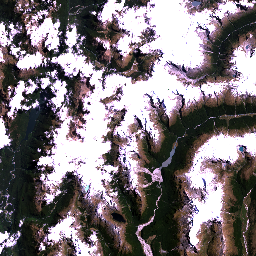}}\\
    \adjustbox{width=\linewidth, frame=0pt 1pt}{\includegraphics{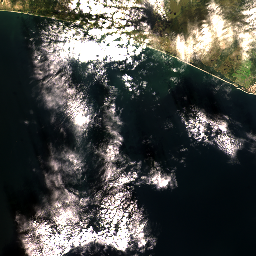}}\\    
    \adjustbox{width=\linewidth, frame=0pt 1pt}{\includegraphics{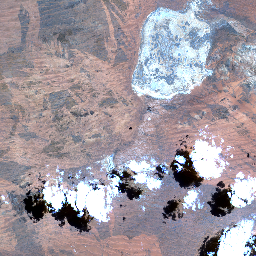}}\\
    \adjustbox{width=\linewidth, frame=0pt 1pt}{\includegraphics{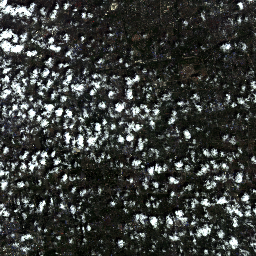}}\\
    \adjustbox{width=\linewidth, frame=0pt 1pt}{\includegraphics{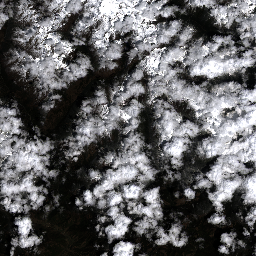}}\\
    \adjustbox{width=\linewidth, frame=0pt 1pt}{\includegraphics{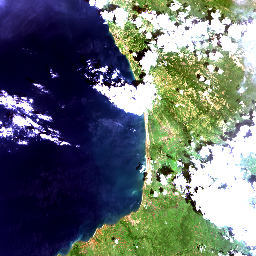}}\\
    \adjustbox{width=\linewidth, frame=0pt 1pt}{\includegraphics{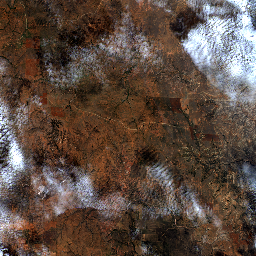}}\\
    \adjustbox{width=\linewidth, frame=0pt 1pt}{\includegraphics{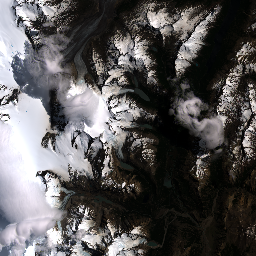}}
\end{subfigure}
\begin{subfigure}{0.1\textwidth}
    \caption*{Ground Truth }
    \adjustbox{width=\linewidth, frame=0pt 1pt}{\includegraphics{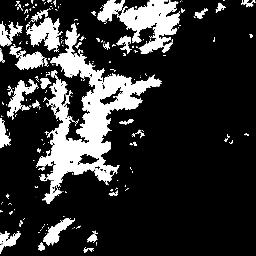}}\\
    \adjustbox{width=\linewidth, frame=0pt 1pt}{\includegraphics{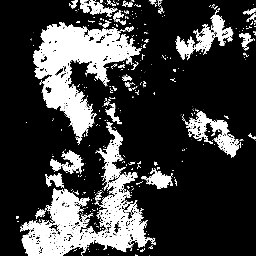}}\\
    \adjustbox{width=\linewidth, frame=0pt 1pt}{\includegraphics{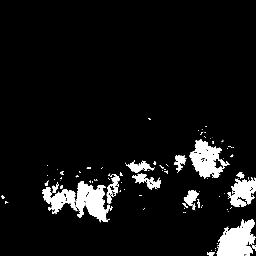}}\\
    \adjustbox{width=\linewidth, frame=0pt 1pt}{\includegraphics{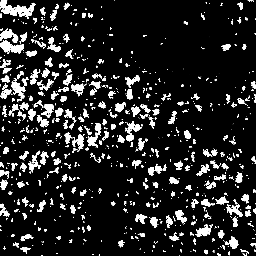}}\\
    \adjustbox{width=\linewidth, frame=0pt 1pt}{\includegraphics{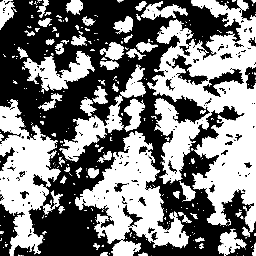}}\\
    \adjustbox{width=\linewidth, frame=0pt 1pt}{\includegraphics{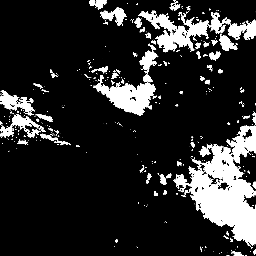}}\\
    \adjustbox{width=\linewidth, frame=0pt 1pt}{\includegraphics{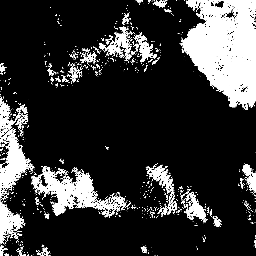}}\\
    \adjustbox{width=\linewidth, frame=0pt 1pt}{\includegraphics{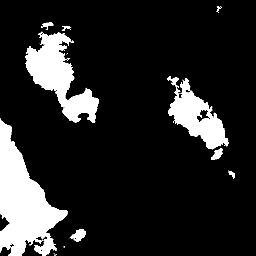}}
\end{subfigure}
\begin{subfigure}{0.1\textwidth}
    \caption*{CDNetV2}
    \adjustbox{width=\linewidth, frame=0pt 1pt}{\includegraphics{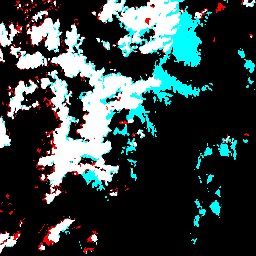}}\\ 
    \adjustbox{width=\linewidth, frame=0pt 1pt}{\includegraphics{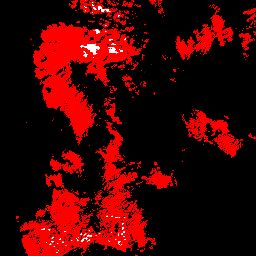}}\\
    \adjustbox{width=\linewidth, frame=0pt 1pt}{\includegraphics{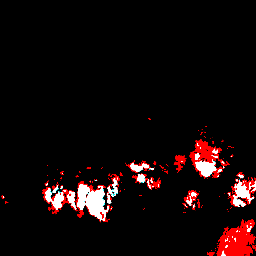}}\\
    \adjustbox{width=\linewidth, frame=0pt 1pt}{\includegraphics{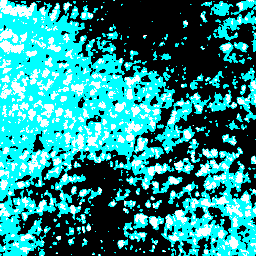}}\\
    \adjustbox{width=\linewidth, frame=0pt 1pt}{\includegraphics{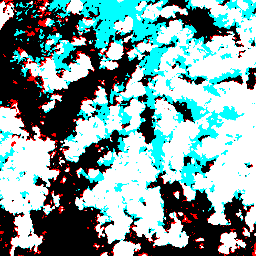}}\\
    \adjustbox{width=\linewidth, frame=0pt 1pt}{\includegraphics{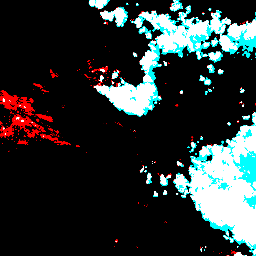}}\\
    \adjustbox{width=\linewidth, frame=0pt 1pt}{\includegraphics{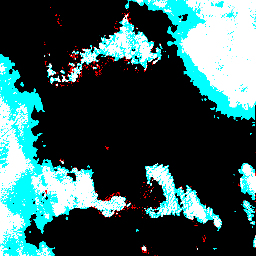}}\\
    \adjustbox{width=\linewidth, frame=0pt 1pt}{\includegraphics{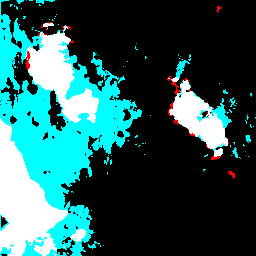}}
\end{subfigure}
\begin{subfigure}{0.1\textwidth}
    \caption*{CDNetV1}
    \adjustbox{width=\linewidth, frame=0pt 1pt}{\includegraphics{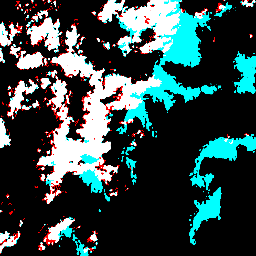}}\\
    \adjustbox{width=\linewidth, frame=0pt 1pt}{\includegraphics{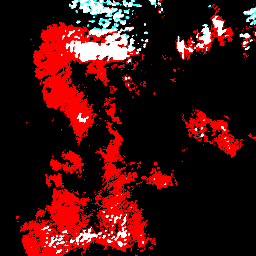}}\\
    \adjustbox{width=\linewidth, frame=0pt 1pt}{\includegraphics{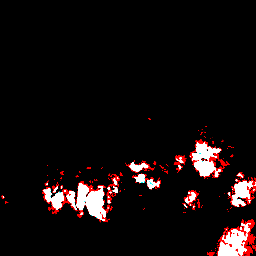}}\\
    \adjustbox{width=\linewidth, frame=0pt 1pt}{\includegraphics{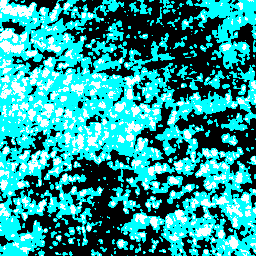}}\\
    \adjustbox{width=\linewidth, frame=0pt 1pt}{\includegraphics{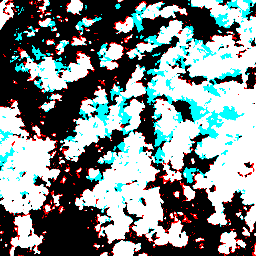}}\\
    \adjustbox{width=\linewidth, frame=0pt 1pt}{\includegraphics{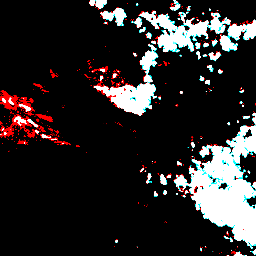}}\\
    \adjustbox{width=\linewidth, frame=0pt 1pt}{\includegraphics{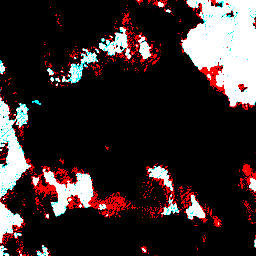}}\\
    \adjustbox{width=\linewidth, frame=0pt 1pt}{\includegraphics{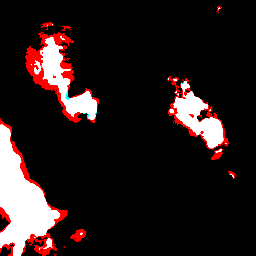}}
\end{subfigure}
\begin{subfigure}{0.1\textwidth}
    \caption*{DeepLabV3+}
    \adjustbox{width=\linewidth, frame=0pt 1pt}{\includegraphics{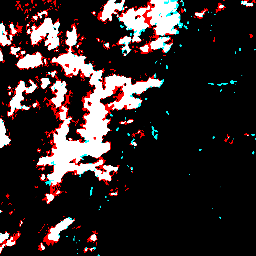}}\\ 
    \adjustbox{width=\linewidth, frame=0pt 1pt}{\includegraphics{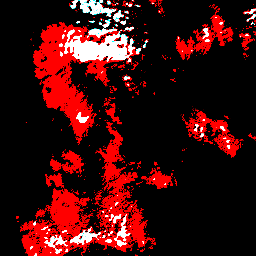}}\\
    \adjustbox{width=\linewidth, frame=0pt 1pt}{\includegraphics{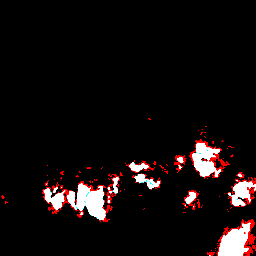}}\\
    \adjustbox{width=\linewidth, frame=0pt 1pt}{\includegraphics{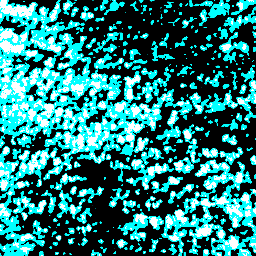}}\\
    \adjustbox{width=\linewidth, frame=0pt 1pt}{\includegraphics{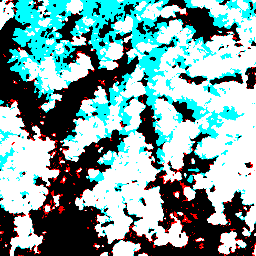}}\\
    \adjustbox{width=\linewidth, frame=0pt 1pt}{\includegraphics{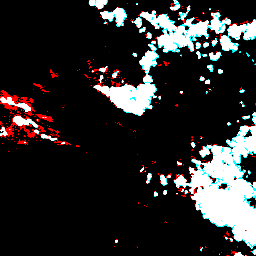}}\\
    \adjustbox{width=\linewidth, frame=0pt 1pt}{\includegraphics{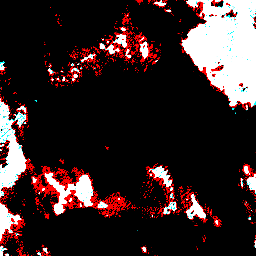}}\\
    \adjustbox{width=\linewidth, frame=0pt 1pt}{\includegraphics{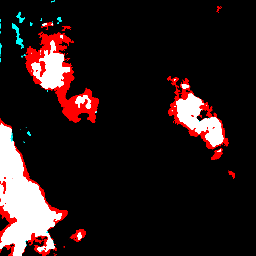}}
\end{subfigure}
\begin{subfigure}{0.1\textwidth}
    \caption*{PSPNet}
    \adjustbox{width=\linewidth, frame=0pt 1pt}{\includegraphics{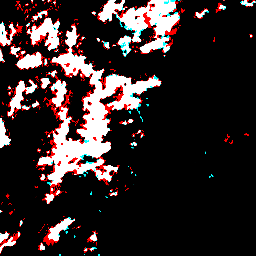}}\\
    \adjustbox{width=\linewidth, frame=0pt 1pt}{\includegraphics{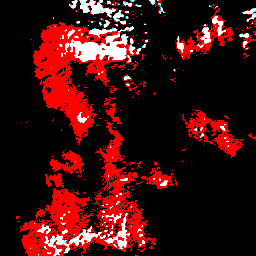}}\\
    \adjustbox{width=\linewidth, frame=0pt 1pt}{\includegraphics{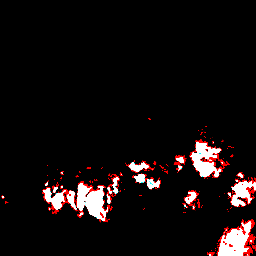}}\\
    \adjustbox{width=\linewidth, frame=0pt 1pt}{\includegraphics{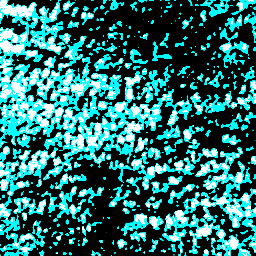}}\\
    \adjustbox{width=\linewidth, frame=0pt 1pt}{\includegraphics{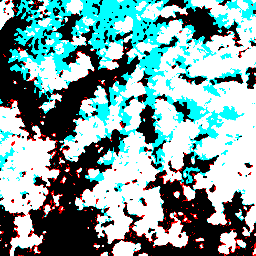}}\\
    \adjustbox{width=\linewidth, frame=0pt 1pt}{\includegraphics{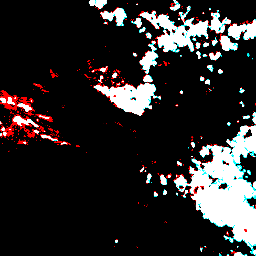}}\\
    \adjustbox{width=\linewidth, frame=0pt 1pt}{\includegraphics{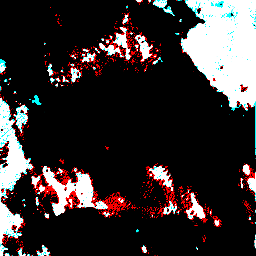}}\\
    \adjustbox{width=\linewidth, frame=0pt 1pt}{\includegraphics{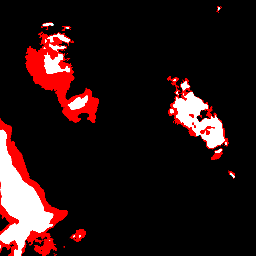}}
\end{subfigure}
\begin{subfigure}{0.1\textwidth}
    \caption*{RS-Net}
    \adjustbox{width=\linewidth, frame=0pt 1pt}{\includegraphics{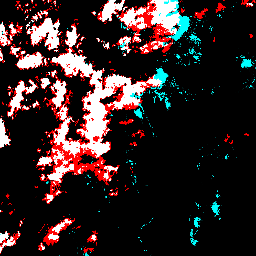}}\\ 
    \adjustbox{width=\linewidth, frame=0pt 1pt}{\includegraphics{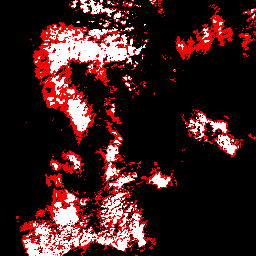}}\\
    \adjustbox{width=\linewidth, frame=0pt 1pt}{\includegraphics{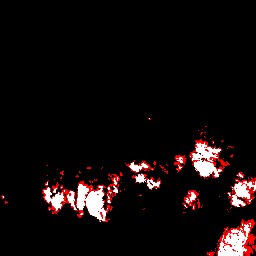}}\\
    \adjustbox{width=\linewidth, frame=0pt 1pt}{\includegraphics{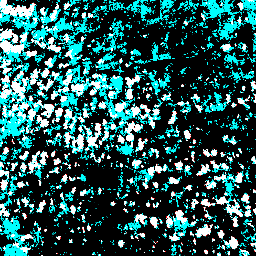}}\\
    \adjustbox{width=\linewidth, frame=0pt 1pt}{\includegraphics{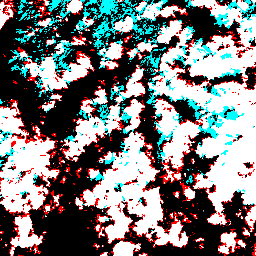}}\\
    \adjustbox{width=\linewidth, frame=0pt 1pt}{\includegraphics{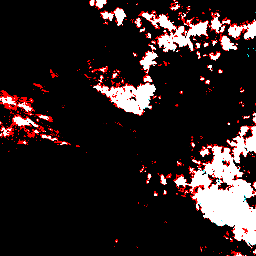}}\\
    \adjustbox{width=\linewidth, frame=0pt 1pt}{\includegraphics{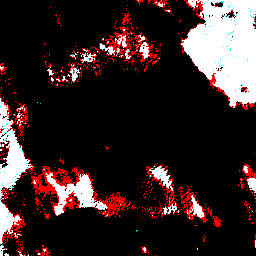}}\\
    \adjustbox{width=\linewidth, frame=0pt 1pt}{\includegraphics{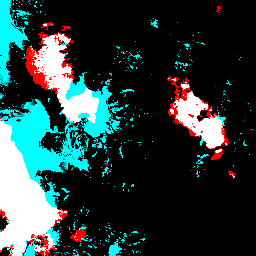}}
\end{subfigure}
\begin{subfigure}{0.1\textwidth}
    \caption*{RefUNetV4}
    \adjustbox{width=\linewidth, frame=0pt 1pt}{\includegraphics{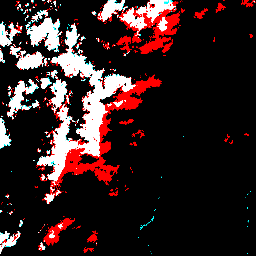}}\\
    \adjustbox{width=\linewidth, frame=0pt 1pt}{\includegraphics{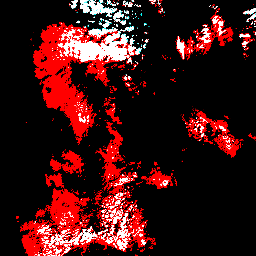}}\\
    \adjustbox{width=\linewidth, frame=0pt 1pt}{\includegraphics{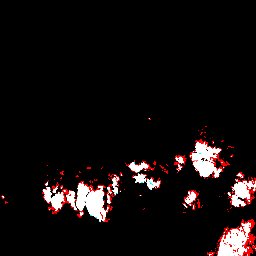}}\\
    \adjustbox{width=\linewidth, frame=0pt 1pt}{\includegraphics{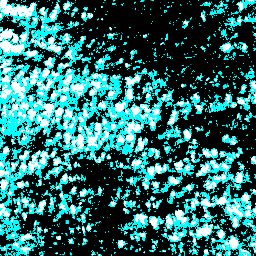}}\\
    \adjustbox{width=\linewidth, frame=0pt 1pt}{\includegraphics{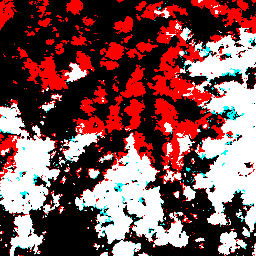}}\\
    \adjustbox{width=\linewidth, frame=0pt 1pt}{\includegraphics{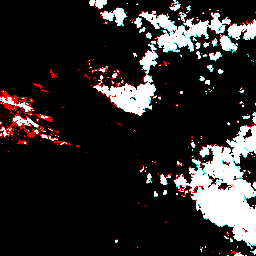}}\\
    \adjustbox{width=\linewidth, frame=0pt 1pt}{\includegraphics{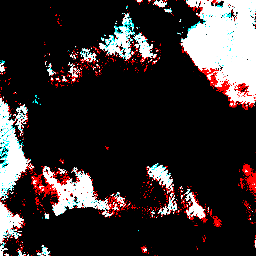}}\\
    \adjustbox{width=\linewidth, frame=0pt 1pt}{\includegraphics{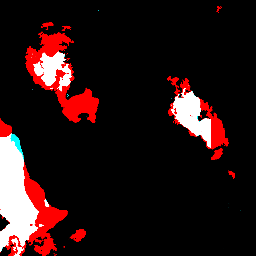}}
\end{subfigure}
\begin{subfigure}{0.1\textwidth}
    \caption*{CLiSA (ours)}
    \adjustbox{width=\linewidth, frame=0pt 1pt}{\includegraphics{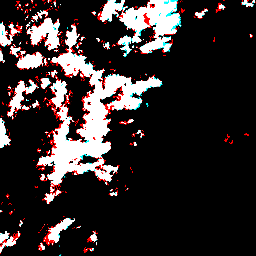}}\\
    \adjustbox{width=\linewidth, frame=0pt 1pt}{\includegraphics{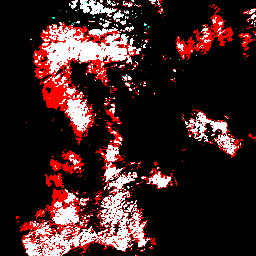}}\\
    \adjustbox{width=\linewidth, frame=0pt 1pt}{\includegraphics{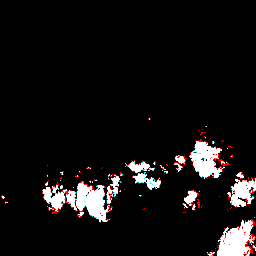}}\\
    \adjustbox{width=\linewidth, frame=0pt 1pt}{\includegraphics{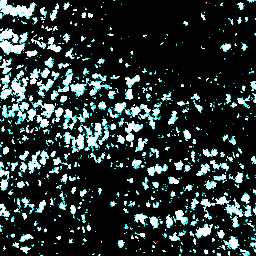}}\\
    \adjustbox{width=\linewidth, frame=0pt 1pt}{\includegraphics{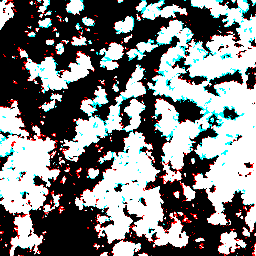}}\\
    \adjustbox{width=\linewidth, frame=0pt 1pt}{\includegraphics{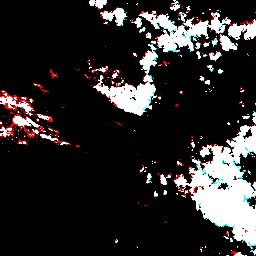}}\\
    \adjustbox{width=\linewidth, frame=0pt 1pt}{\includegraphics{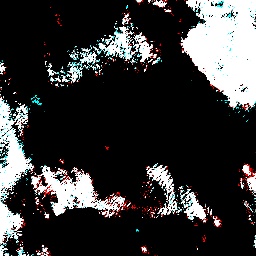}}\\
    \adjustbox{width=\linewidth, frame=0pt 1pt}{\includegraphics{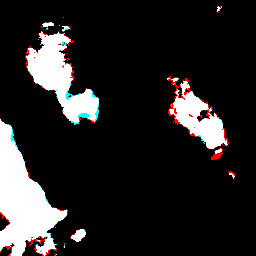}}
\end{subfigure}
Legend \quad \tikz \fill [red] rectangle (1, 0.2); Omission error \quad
\tikz \fill [Mycolor2] rectangle (1, 0.2); Commission error
\caption{Comparison of cloud mask generation results on Biome dataset with different signatures. For each model the best output corresponding to the band combination $(C)$ is shown.}
\label{fig:fig1_results}
\end{figure*}
\section{Experiments}\label{exp}
In this section, we discuss experiments to analyze the efficacy of our proposed model for cloud segmentation.
\subsection{Datasets and Study Area}
We use publicly available Biome dataset \cite{4005} consisting of 96 Landsat-8 scenes with different signatures and 11 spectral bands with effective ground sampling distance (GSD) of 30m for training our cloud segmentation model. To visualize the effect of different bands in generated cloud-segmented maps, we have carried out experiments with different band combinations. In regards to this, we have generated a paired dataset consisting of 70,000 ground truth and input patches with size $(256,256)$ and $(256,256, C)$, respectively. Here $C\in \{1,3,4,11\}$ corresponds to different band combinations of G, RGB, RGB+NIR, and all bands respectively. We use 30,000 samples for training, 20,000 for validation, and 20,000 for testing. Apart from this we also cross-validated our model on other Landsat-8 dataset like SPARCS \cite{4006} consisting of 80 paired patches of size (1000, 1000) and other satellite datasets also through transfer learning. This includes Sentinel-2 (GSD=10m) cloud mask catalog \cite{4008} with 520 patches of size (1022,1022) and our self-crafted dataset from 50 scenes of high-resolution multi-spectral payload of Cartosat-2S satellite (GSD=1.6m). In both cases, transfer learning is performed for 2000 extracted patches of shape (256,256), and due to variation of band definition, it is only performed for $C\in \{1,3,4\}$ band combination.

\subsection{Implementation Details}
All the experiments are carried out in identical environment. We use an ADAM optimizer with an exponentially varying learning rate of 0.0001. During adversarial training, we update the critic once every single update in the generator. We set $\lambda_{CE}=10$, $\lambda_{IoU}=0.8$, $\lambda_{ADV}=0.1$, and $\lambda_{L2}=0.01$. Among other hyperparameters, we set the batch size as 64 and a maximum number of training iterations as 100. The entire framework is developed using PyTorch. All the experiments are conducted in 2 Nvidia V100 with $40$GB GPU memory.
\section{Results and analysis}\label{res}
In this section, we analyze both qualitatively and quantitatively the performance of our proposed model. We compare it with other state-of-the-art methods in cloud segmentation research area like CDNetV1 \cite{19}, CDNetV2 \cite{20}, RS-Net \cite{18} and RefUNetV4 \cite{4010} . In addition, we also use two representative segmentation modules, namely PSPnet \cite{22}, and, DeeplabV3+ \cite{23} for our comparative study. Along with traditional performance metrics like overall accuracy (OA), recall (Rec), precision (Pre), and, mean IoU (mIoU), we leverage the usage of kappa coefficient (Kappa), boundary IoU (BI) \cite{25}, and, hausdraff distance (HD) for comprehensive analysis of our model performance.
    
    

\begin{table*}[]
\caption{Quantitative analysis on SPARCS dataset and comparison with other state-of-the-art methods. First and second methods are highlighted in red and green, respectively.}
\label{tab:my-table2}
\resizebox{\textwidth}{!}{%
\begin{tabular}{c|ccccccc|ccccccc}
\hline
Band Configuration & \multicolumn{7}{c|}{C=1}                                                                                                                                                                                                                                                                & \multicolumn{7}{c}{C=3}                                                                                                                                                                                                                                                                 \\ \hline
Method             & OA                                    & Rec                                   & Pre                                   & mIoU(\%)                              & Kappa                                 & BI(\%)                                & HD (m)                                  & OA                                    & Rec                                   & Pre                                   & mIoU(\%)                              & Kappa                                 & BI(\%)                                & HD (m)                                  \\ \hline
CDNetV2 \cite{20}          & 90.88                                 & 81.77                                 & 87.56                                 & 84.52                                 & 68.53                                 & 23.18                                 & 4732.56                                 & {\color[HTML]{009901} \textit{92.07}} & 82.83                                 & 90.65                                 & 86.07                                 & 72.12                                 & 26.62                                 & 3375.12                                 \\
CDNetV1 \cite{19}           & 90.21                                 & 80.43                                 & 86.49                                 & 83.83                                 & 66.02                                 & 25.76                                 & 5009.07                                 & 90.15                                 & {\color[HTML]{009901} \textit{85.84}} & 83.74                                 & 84.29                                 & 69.46                                 & 30.63                                 & 4233.48                                 \\
DeepLabV3+ \cite{23}        & 87.79                                 & 77.1                                  & 81.57                                 & 80.09                                 & 58.1                                  & 22.40                                 & 5546.19                                 & 85.88                                 & 79.61                                 & 77.29                                 & 78.86                                 & 56.72                                 & 28.04                                 & 4959.69                                 \\
PSPnet \cite{22}            & 82.62                                 & 75.9                                  & 72.73                                 & 75.28                                 & 48.22                                 & {  18.92}          & 4832.57                                 & 90.99                                 & 81.47                                 & 88.14                                 & 84.93                                 & 68.58                                 & 26.69                                 & 4173.38                                 \\
RS-Net \cite{18}            & 81.65                                 & 73.16                                 & {  72.18}          & 73.57                                 & 40.39                                 & 20.68                                 & 6758.24                                 & 83.73                                 & 70.09                                 & 74.18                                 & 74.68                                 & 43.65                                 & 23.85                                 & 4219.27                                 \\
RefUNetV4 \cite{4010}           & {\color[HTML]{009901} \textit{91.65}} & {\color[HTML]{009901} \textit{82.24}} & {\color[HTML]{9A0000} \textbf{92.83}} & {\color[HTML]{009901} \textit{86.22}} & {\color[HTML]{009901} \textit{72.29}} & {\color[HTML]{009901} \textit{33.46}} & {\color[HTML]{009901} \textit{3965.76}} & {  92.03}          & {  83.37}          & {\color[HTML]{009901} \textit{91.42}} & {\color[HTML]{009901} \textit{86.84}} & {\color[HTML]{009901} \textit{73.37}} & {\color[HTML]{009901} \textit{34.08}} & {\color[HTML]{009901} \textit{3526.44}} \\
CLiSA (ours)               & {\color[HTML]{9A0000} \textbf{93.85}} & {\color[HTML]{9A0000} \textbf{86.09}} & {\color[HTML]{009901} \textit{90.56}} & {\color[HTML]{9A0000} \textbf{89.13}} & {\color[HTML]{9A0000} \textbf{73.53}} & {\color[HTML]{9A0000} \textbf{58.26}} & {\color[HTML]{9A0000} \textbf{1945.53}} & {\color[HTML]{9A0000} \textbf{95.52}} & {\color[HTML]{9A0000} \textbf{86.14}} & {\color[HTML]{9A0000} \textbf{94.79}} & {\color[HTML]{9A0000} \textbf{91.48}} & {\color[HTML]{9A0000} \textbf{78.55}} & {\color[HTML]{9A0000} \textbf{60.27}} & {\color[HTML]{9A0000} \textbf{2182.82}} \\ \hline
Band Configuration & \multicolumn{7}{c|}{C=4}                                                                                                                                                                                                                                                                & \multicolumn{7}{c}{C=11}                                                                                                                                                                                                                                                                \\ \hline
Method             & OA                                    & Rec                                   & Pre                                   & mIoU(\%)                              & Kappa                                 & BI(\%)                                & HD (m)                                  & OA                                    & Rec                                   & Pre                                   & mIoU(\%)                              & Kappa                                 & BI(\%)                                & HD (m)                                  \\ \hline
CDNetV2 \cite{20}         & 89.51                                 & 86.74                                 & 82.42                                 & 82.89                                 & 68.64                                 & 23.27                                 & 4248.81                                 & 88.42                                 & 70.35                                 & 92.95                                 & 80.78                                 & 52.34                                 & 18.72                                 & 8187.98                                 \\
CDNetV1 \cite{19}           & 91.88                                 & 83.95                                 & 88.95                                 & 86.1                                  & 72.32                                 & 31.03                                 & 4581.81                                 & 92.77                                 & {\color[HTML]{009901} \textit{85.11}} & 90.85                                 & 87.28                                 & 75.23                                 & 29.12                                 & 3856.01                                 \\
DeepLabV3+ \cite{23}        & 89.30                                 & 85.09                                 & 82.33                                 & 82.72                                 & 67.21                                 & 30.7                                  & 4549.95                                 & 92.58                                 & 84.71                                 & 90.57                                 & 86.95                                 & 74.52                                 & 28.20                                 & 3440.02                                 \\
PSPnet \cite{22}            & {\color[HTML]{009901} \textit{93.19}} & {\color[HTML]{009901} \textit{85.73}} & {\color[HTML]{9A0000} \textbf{91.68}}          & {\color[HTML]{009901} \textit{87.83}} & {\color[HTML]{9A0000} \textbf{76.63}} & {  31.36}          & 3593.78                                 & 92.94                                 & {  84.44}          & 92.17                                 & 87.59                                 & 75.3                                  & 31.13                                 & 3269.02                                 \\
RS-Net \cite{18}            & 85.24                                 & 71.92                                 & {  77.24}          & 76.58                                 & 48.23                                 & 27.42                                 & 3844.34                                 & 85.41                                 & 70.35                                 & {  78.38}          & 76.58                                 & {  46.67}          & 27.79                                 & 4029.91                                 \\
RefUNetV4 \cite{4010}           & {  92.24}          & {  83.55}          & {\color[HTML]{009901} \textit{90.54}} & {  86.67}          & {  73.14}          & {\color[HTML]{009901} \textit{32.89}} & {\color[HTML]{009901} \textit{3184.41}} & {\color[HTML]{009901} \textit{93.26}} & {  84.42}          & {\color[HTML]{9A0000} \textbf{93.42}} & {\color[HTML]{009901} \textit{88.16}} & {\color[HTML]{009901} \textit{76.15}} & {\color[HTML]{009901} \textit{35.58}} & {\color[HTML]{009901} \textit{3464.93}} \\
CLiSA (ours)               & {\color[HTML]{9A0000} \textbf{94.06}} & {\color[HTML]{9A0000} \textbf{87.61}} & {  87.51}          & {\color[HTML]{9A0000} \textbf{89.08}} & {\color[HTML]{009901} \textit{76.06}} & {\color[HTML]{9A0000} \textbf{62.64}} & {\color[HTML]{9A0000} \textbf{1901.49}} & {\color[HTML]{9A0000} \textbf{96.06}} & {\color[HTML]{9A0000} \textbf{89.4}}  & {\color[HTML]{009901} \textit{93.02}} & {\color[HTML]{9A0000} \textbf{91.74}} & {\color[HTML]{9A0000} \textbf{81.22}} & {\color[HTML]{9A0000} \textbf{67.24}} & {\color[HTML]{9A0000} \textbf{1713.19}} \\ \hline
\end{tabular}%
}
\end{table*}

\begin{table*}[]
\caption{Quantitative analysis on Sentinel-2 dataset and comparison with other state-of-the-art methods. First and second methods are highlighted in red and green, respectively.}
\label{tab:my-table3}
\resizebox{\textwidth}{!}{%
\begin{tabular}{c|ccccccc|ccccccc|ccccccc}
\hline
Band Configuration & \multicolumn{7}{c|}{C=1}                                                                                                                                                                                                                                                                & \multicolumn{7}{c|}{C=3}                                                                                                                                                                                                                                                                & \multicolumn{7}{c}{C=4}                                                                                                                                                                                                                                                                 \\ \hline
Method             & OA                                    & Rec                                   & Pre                                   & mIoU(\%)                              & Kappa                                 & BI(\%)                                & HD (m)                                  & OA                                    & Rec                                   & Pre                                   & mIoU(\%)                              & Kappa                                 & BI(\%)                                & HD (m)                                  & OA                                    & Rec                                   & Pre                                   & mIoU(\%)                              & Kappa                                 & BI(\%)                                & HD (m)                                  \\ \hline
CDNetV2 \cite{20}          & 70.87                                 & 70.63                                 & 74.96                                 & 65.82                                 & 40.16                                 & 22.59                                 & 7832.01                                 & {  75.88}          & 80.33                                 & 72.73                                 & 72.99                                 & 62.15                                 & 28.75                                 & 3418.31                                 & {  72.96}          & 72.71                                 & 74.13                                 & 70.98                                 & 60.77                                 & 27.15                                 & 3230.33                                 \\
CDNetV1 \cite{19}           & 68.07                                 & 70.49                                 & 67.61                                 & 61.94                                 & 38.59                                 & 20.76                                 & 8042.26                                 & 72.42                                 & {  72.29}          & 72.75                                 & 65.75                                 & 50.46                                 & 23.90                                 & 4264.91                                 & 75.17                                 & {  75.09}          & 75.28                                 & 69.47                                 & 65.43                                 & 30.63                                 & 2822.64                                 \\
DeepLabV3+ \cite{23}        & {\color[HTML]{009901} \textit{76.13}} & 75.82                                 & {\color[HTML]{9A0000} \textbf{77.81}} & {\color[HTML]{009901} \textit{70.34}} & {\color[HTML]{009901} \textit{60.05}} & 30.13                                 & 3426.24                                 & 82.27                                 & 82.43                                 & 82.67                                 & 77.19                                 & {\color[HTML]{009901} \textit{72.61}} & 36.31                                 & 1737.82                                 & 81.03                                 & {\color[HTML]{009901} \textit{82.17}} & 80.36                                 & {\color[HTML]{009901} \textit{78.82}} & {\color[HTML]{009901} \textit{74.11}} & 35.98                                 & 1840.83                                 \\
PSPnet \cite{22}            & 66.13                                 & 65.93                                 & 60.73                                 & 54.64                                 & 44.11                                 & {  18.54}          & 9639.16                                 & 67.05                                 & 62.81                                 & 67.30                                 & 56.97                                 & 49.97                                 & 21.30                                 & 6449.72                                 & 69.05                                 & 67.67                                 & 68.44                                 & 60.80                                 & 54.96                                 & 21.11                                 & 4347.15                                 \\
RS-Net \cite{18}            & 71.24                                 & {\color[HTML]{009901} \textit{76.34}} & {  69.46}          & 65.41                                 & 51.26                                 & {\color[HTML]{009901} \textit{34.24}} & {\color[HTML]{009901} \textit{3008.45}} & {\color[HTML]{009901} \textit{83.96}} & {\color[HTML]{009901} \textit{84.81}} & {\color[HTML]{009901} \textit{85.35}} & {\color[HTML]{009901} \textit{80.15}} & 71.83                                 & {\color[HTML]{009901} \textit{39.27}} & {\color[HTML]{009901} \textit{1528.94}} & {\color[HTML]{009901} \textit{81.68}} & 81.43                                 & {\color[HTML]{009901} \textit{82.68}} & 75.75                                 & 70.19                                 & {\color[HTML]{009901} \textit{40.34}} & {\color[HTML]{009901} \textit{1678.76}} \\
RefUNetV4 \cite{4010}           & {  68.45}          & {  72.07}          & {  67.96}          & {  62.83}          & {  42.45}          & {  28.25}          & {  3399.35}          & {  73.9}           & {  74.72}          & {  75.56}          & {  71.78}          & {  61.51}          & {  32.51}          & {  2652.54}          & {  78.93}          & {  78.80}          & {  79.35}          & {  73.56}          & {  66.81}          & {  35.53}          & {  2289.32}          \\
CLiSA (ours)               & {\color[HTML]{9A0000} \textbf{81.78}} & {\color[HTML]{9A0000} \textbf{82.31}} & {\color[HTML]{009901} \textit{76.24}} & {\color[HTML]{9A0000} \textbf{78.35}} & {\color[HTML]{9A0000} \textbf{66.16}} & {\color[HTML]{9A0000} \textbf{42.02}} & {\color[HTML]{9A0000} \textbf{2618.66}} & {\color[HTML]{9A0000} \textbf{89.3}}  & {\color[HTML]{9A0000} \textbf{88.07}} & {\color[HTML]{9A0000} \textbf{88.31}} & {\color[HTML]{9A0000} \textbf{86.26}} & {\color[HTML]{9A0000} \textbf{79.77}} & {\color[HTML]{9A0000} \textbf{45.51}} & {\color[HTML]{9A0000} \textbf{1227.11}} & {\color[HTML]{9A0000} \textbf{88.42}} & {\color[HTML]{9A0000} \textbf{88.33}} & {\color[HTML]{9A0000} \textbf{89.6}}  & {\color[HTML]{9A0000} \textbf{85.83}} & {\color[HTML]{9A0000} \textbf{80.11}} & {\color[HTML]{9A0000} \textbf{44.62}} & {\color[HTML]{9A0000} \textbf{1360.28}} \\ \hline
\end{tabular}%
}
\end{table*}
\begin{table*}[]
\caption{Quantitative analysis on Cartosat-2S dataset and comparison with other state-of-the-art methods. First and second methods are highlighted in red and green, respectively.}
\label{tab:my-table4}
\resizebox{\textwidth}{!}{%
\begin{tabular}{c|ccccccc|ccccccc|ccccccc}
\hline
Band Configuration & \multicolumn{7}{c|}{C=1}                                                                                                                                                                                                                                                                & \multicolumn{7}{c|}{C=3}                                                                                                                                                                                                                                                               & \multicolumn{7}{c}{C=4}                                                                                                                                                                                                                                                                 \\ \hline
Method             & OA                                    & Rec                                   & Pre                                   & mIoU(\%)                              & Kappa                                 & BI(\%)                                & HD (m)                                  & OA                                    & Rec                                   & Pre                                   & mIoU(\%)                              & Kappa                                 & BI(\%)                                & HD (m)                                 & OA                                    & Rec                                   & Pre                                   & mIoU(\%)                              & Kappa                                 & BI(\%)                                & HD (m)                                  \\ \hline
CDNetV2 \cite{20}           & 89.27                                 & 88.69                                 & 87.87                                 & 85.61                                 & 79.29                                 & 48.46                                 & 2546.45                                 & {90.59}          & 91.40                                 & 90.02                                 & 85.46                                 & 80.15                                 & 50.29                                 & 2106.49                                & {87.51}          & 89.48                                 & 85.27                                 & 82.54                                 & 75.16                                 & 47.59                                 & 1964.67                                 \\
CDNetV1 \cite{19}           & 89.98                                 & 89.04                                 & 88.68                                 & 83.34                                 & 75.48                                 & 46.78                                 & 2354.52                                 & 91.16                                 & {  91.07}          & 91.64                                 & 84.78                                 & 78.66                                 & 52.78                                 & 1884.27                                & 87.08                                 & {  87.26}          & 87.78                                 & 84.04                                 & 74.69                                 & 48.12                                 & 1807.96                                 \\
DeepLabV3+ \cite{23}        & {85.49}          & 88.12                                 & {82.77}          & {80.94}          & {72.65}          & 44.49                                 & 2816.75                                 & 92.86                                 & 92.22                                 & 92.68                                 & 88.59                                 & {  85.41}          & 50.05                                 & 1506.01                                & 89.38                                 & {88.14}          & 88.41                                 & {88.06}          & {79.66}          & 52.73                                 & 1667.15                                 \\
PSPnet \cite{22}             & 92.44                                 & 93.08                                 & 93.26                                 & 88.42                                 & 81.49                                 & {50.49}          & 1806.87                                 & 87.88                                 & 89.92                                 & 88.40                                 & 80.19                                 & 76.64                                 & 43.34                                 & 2333.27                                & 85.84                                 & 86.62                                 & 84.18                                 & 79.26                                 & 72.76                                 & 45.57                                 & 2106.35                                 \\
RS-Net \cite{18}            & 93.78                                 & {  91.49}          & {\color[HTML]{009901} \textit{94.07}} & 91.08                                 & {\color[HTML]{009901} \textit{86.49}} & {  54.24}          & {  1550.87}          & {  93.81}          & {  91.54}          & {\color[HTML]{009901} \textit{95.78}} & {  90.15}          & 84.86                                 & {  62.27}          & {  1048.15}         & {  91.22}          & 90.04                                 & {  90.23}          & {\color[HTML]{009901} \textit{89.48}} & 82.24                                 & {  56.48}          & {  1870.26}          \\
RefUNetV4 \cite{4010}           & {\color[HTML]{009901} \textit{94.29}} & {\color[HTML]{680100} \textbf{96.08}} & {  92.00}          & {\color[HTML]{009901} \textit{91.46}} & {  85.78}          & {\color[HTML]{009901} \textit{60.46}} & {\color[HTML]{009901} \textit{1405.47}} & {\color[HTML]{009901} \textit{96.03}} & {\color[HTML]{009901} \textit{95.27}} & {  95.49}          & {\color[HTML]{009901} \textit{93.46}} & {\color[HTML]{009901} \textit{90.49}} & {\color[HTML]{009901} \textit{69.57}} & {\color[HTML]{009901} \textit{675.12}} & {\color[HTML]{009901} \textit{91.47}} & {\color[HTML]{009901} \textit{90.80}} & {\color[HTML]{009901} \textit{91.78}} & {  89.24}          & {\color[HTML]{009901} \textit{87.46}} & {\color[HTML]{009901} \textit{61.25}} & {\color[HTML]{009901} \textit{1589.27}} \\
CLiSA (ours)               & {\color[HTML]{680100} \textbf{96.82}} & {\color[HTML]{009901} \textit{95.40}} & {\color[HTML]{680100} \textbf{94.74}} & {\color[HTML]{680100} \textbf{95.22}} & {\color[HTML]{680100} \textbf{93.25}} & {\color[HTML]{680100} \textbf{75.51}} & {\color[HTML]{680100} \textbf{540.24}}  & {\color[HTML]{9A0000} \textbf{97.90}} & {\color[HTML]{9A0000} \textbf{96.35}} & {\color[HTML]{9A0000} \textbf{96.16}} & {\color[HTML]{9A0000} \textbf{96.33}} & {\color[HTML]{9A0000} \textbf{94.76}} & {\color[HTML]{9A0000} \textbf{78.54}} & {\color[HTML]{9A0000} \textbf{354.26}} & {\color[HTML]{9A0000} \textbf{95.28}} & {\color[HTML]{9A0000} \textbf{92.33}} & {\color[HTML]{9A0000} \textbf{96.58}} & {\color[HTML]{9A0000} \textbf{94.26}} & {\color[HTML]{9A0000} \textbf{90.11}} & {\color[HTML]{9A0000} \textbf{72.56}} & {\color[HTML]{9A0000} \textbf{781.67}}  \\ \hline
\end{tabular}%
}
\end{table*}
\subsection{Analysis on Biome dataset}
\subsubsection{Quantitative analysis}
Table \ref{tab:my-table} visualizes a quantitative comparison of the Biome dataset for the outcomes of our cloud mask generation model with other benchmark techniques. As shown in the table, our model outperforms others in terms of the mentioned evaluation metrics for all band combinations. Compared to the second best method, our model not only provides 2-3\% improvement in OA, and mIoU, but also achieves performance gain by more than 3-6\% for Kappa, 20-40\% in BI, and, 35-55\% in HD. This means our model can generate an accurate cloud mask closer to the ground truth with higher confidence and lesser spatial dissimilarities in the position and shape of segmented regions. Even with only green band configuration ($C=1$), our model excels in generating precise cloud masks compared to higher band combinations $(C=\{3,4,11\})$ for other models. Hence, our proposed model exhibits robust performance irrespective of band configuration due to incorporating dense features of the deeper network to initial layers with HC2A. This observation becomes more obvious from Figure \ref{fig:signature} where we compare the performance of each model in the presence of different signatures. From this figure, it is apparent that our model surpasses others in terms of performance in complex scenarios like snow and water bodies for all band combinations. This observation is significant as for complex situations like snow-cloud coexistence, other model performs preferable cloud segmentation for specific band combination for which it is relatively simple to distinguish between these signatures. However, due to multi-level cross-attention along with spatial and channel self-attention, our model can perform more robustly despite any band combination.

\begin{figure*}
\centering
\begin{subfigure}{0.009\linewidth}
    \caption*{a.}
    \vspace{1.4cm}
    \caption*{b.}
    \vspace{1.4cm}
    \caption*{c.}
    \vspace{1.4cm}
    \caption*{d.}
    \vspace{1.24cm}
    \caption*{e.}
    \vspace{1.28cm}
    \caption*{f.}
    \vspace{1.3cm}
    \caption*{g.}
    \vspace{1.38cm}
    \caption*{h.}
    \vspace{1.4cm}
\end{subfigure}
\begin{subfigure}{0.1\linewidth}
    \caption*{Input Image} 
    \adjustbox{width=\linewidth, frame=0pt 1pt}{\includegraphics{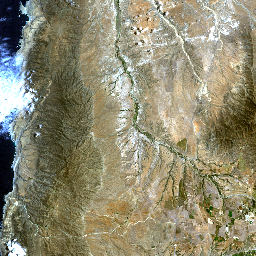}}\\
    \adjustbox{width=\linewidth, frame=0pt 1pt}{\includegraphics
    {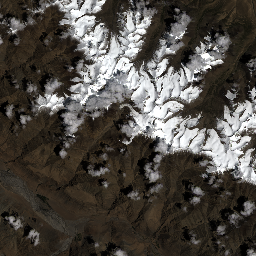}}\\
    \adjustbox{width=\linewidth, frame=0pt 1pt}{\includegraphics{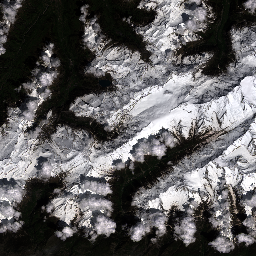}}\\
    \adjustbox{width=\linewidth, frame=0pt 1pt}{\includegraphics{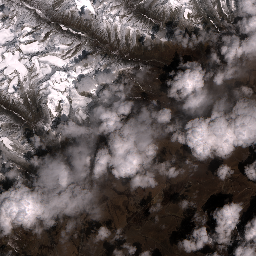}}\\
    \adjustbox{width=\linewidth, frame=0pt 1pt}{\includegraphics{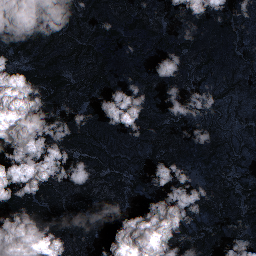}}\\
    \adjustbox{width=\linewidth, frame=0pt 1pt}{\includegraphics{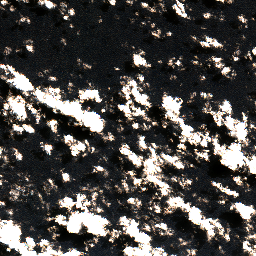}}\\
    \adjustbox{width=\linewidth, frame=0pt 1pt}{\includegraphics{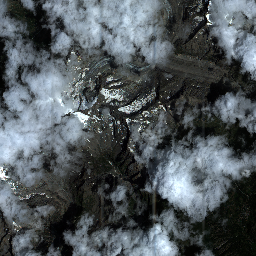}}\\
    \adjustbox{width=\linewidth, frame=0pt 1pt}{\includegraphics{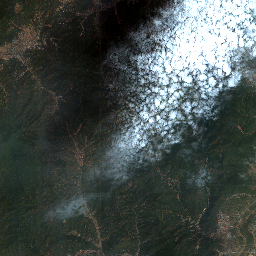}}
\end{subfigure}
\begin{subfigure}{0.1\textwidth}
    \caption*{Ground Truth }
    \adjustbox{width=\linewidth, frame=0pt 1pt}{\includegraphics{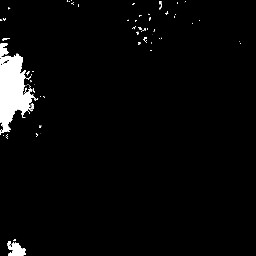}}\\
    \adjustbox{width=\linewidth, frame=0pt 1pt}{\includegraphics{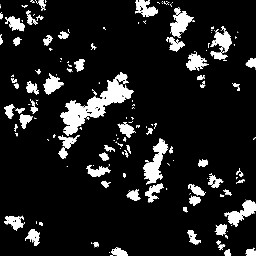}}\\
    \adjustbox{width=\linewidth, frame=0pt 1pt}{\includegraphics{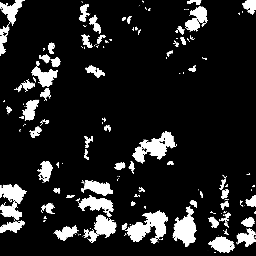}}\\
    \adjustbox{width=\linewidth, frame=0pt 1pt}{\includegraphics{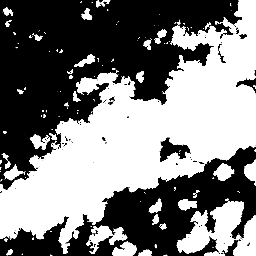}}\\
    \adjustbox{width=\linewidth, frame=0pt 1pt}{\includegraphics{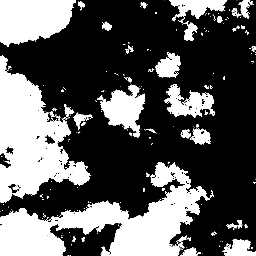}}\\
    \adjustbox{width=\linewidth, frame=0pt 1pt}{\includegraphics{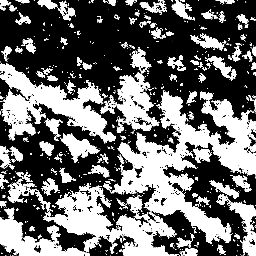}}\\
    \adjustbox{width=\linewidth, frame=0pt 1pt}{\includegraphics{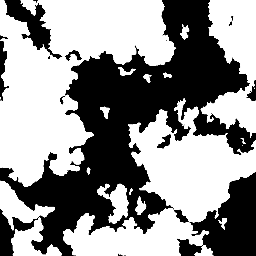}}\\
    \adjustbox{width=\linewidth, frame=0pt 1pt}{\includegraphics{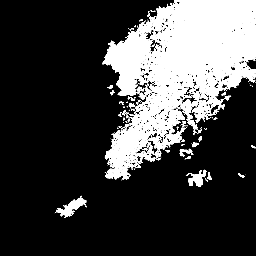}}
\end{subfigure}
\begin{subfigure}{0.1\textwidth}
    \caption*{CDNetV2}
    \adjustbox{width=\linewidth, frame=0pt 1pt}{\includegraphics{other_visuals/sparc_comparison1/cdnetv2_LC80010812013365LGN00_cloud_mask_RGB.png}}\\
    \adjustbox{width=\linewidth, frame=0pt 1pt}{\includegraphics{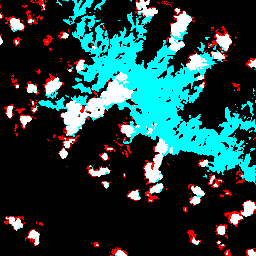}}\\
    \adjustbox{width=\linewidth, frame=0pt 1pt}{\includegraphics{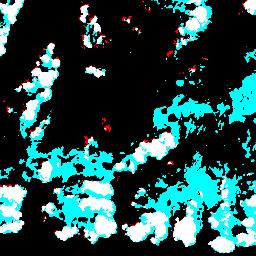}}\\
    \adjustbox{width=\linewidth, frame=0pt 1pt}{\includegraphics{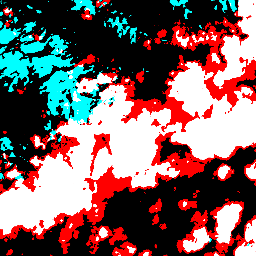}}\\
    \adjustbox{width=\linewidth, frame=0pt 1pt}{\includegraphics{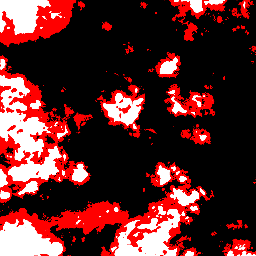}}\\
    \adjustbox{width=\linewidth, frame=0pt 1pt}{\includegraphics{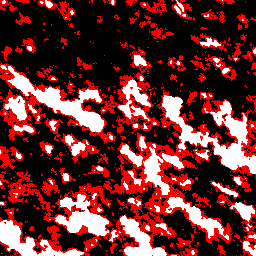}}\\
    \adjustbox{width=\linewidth, frame=0pt 1pt}{\includegraphics{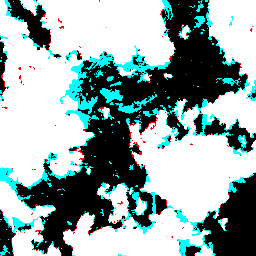}}\\
    \adjustbox{width=\linewidth, frame=0pt 1pt}{\includegraphics{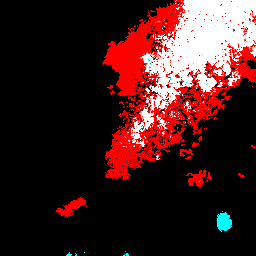}}
    \end{subfigure}
\begin{subfigure}{0.1\textwidth}
    \caption*{CDNetV1}
    \adjustbox{width=\linewidth, frame=0pt 1pt}{\includegraphics{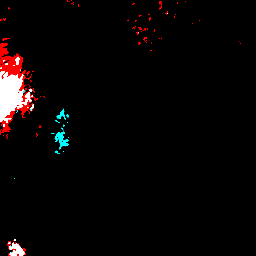}}\\
    \adjustbox{width=\linewidth, frame=0pt 1pt}{\includegraphics{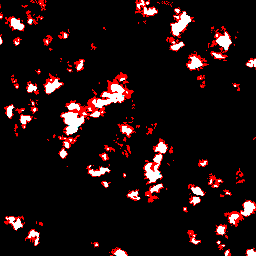}}\\
    \adjustbox{width=\linewidth, frame=0pt 1pt}{\includegraphics{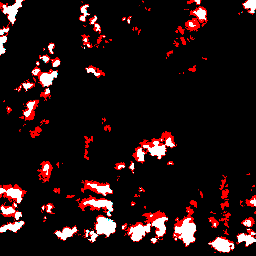}}\\
    \adjustbox{width=\linewidth, frame=0pt 1pt}{\includegraphics{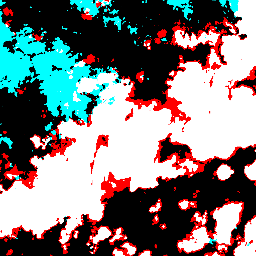}}\\
    \adjustbox{width=\linewidth, frame=0pt 1pt}{\includegraphics{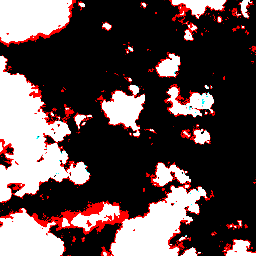}}\\
    \adjustbox{width=\linewidth, frame=0pt 1pt}{\includegraphics{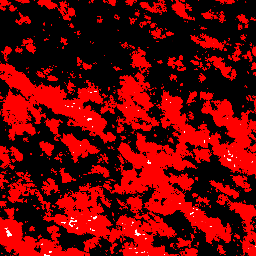}}\\
    \adjustbox{width=\linewidth, frame=0pt 1pt}{\includegraphics{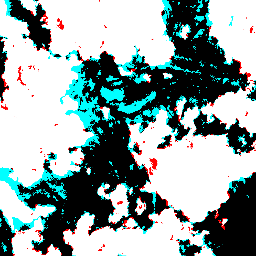}}\\
    \adjustbox{width=\linewidth, frame=0pt 1pt}{\includegraphics{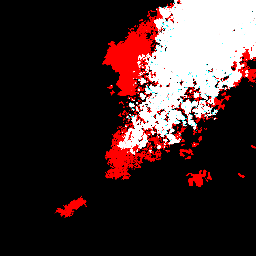}}
    \end{subfigure}
\begin{subfigure}{0.1\textwidth}
    \caption*{DeepLabV3+}
    \adjustbox{width=\linewidth, frame=0pt 1pt}{\includegraphics{other_visuals/sparc_comparison1/deeplabv3+_LC80010812013365LGN00_cloud_mask_RGB.png}}\\
    \adjustbox{width=\linewidth, frame=0pt 1pt}{\includegraphics{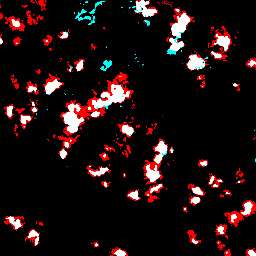}}\\
    \adjustbox{width=\linewidth, frame=0pt 1pt}{\includegraphics{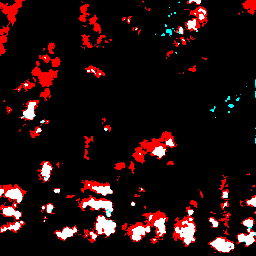}}\\
    \adjustbox{width=\linewidth, frame=0pt 1pt}{\includegraphics{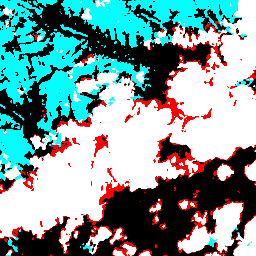}}\\
    \adjustbox{width=\linewidth, frame=0pt 1pt}{\includegraphics{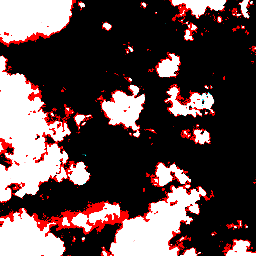}}\\
    \adjustbox{width=\linewidth, frame=0pt 1pt}{\includegraphics{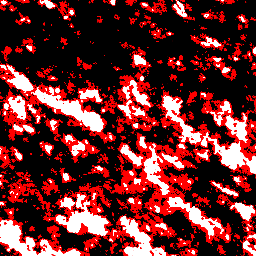}}\\
    \adjustbox{width=\linewidth, frame=0pt 1pt}{\includegraphics{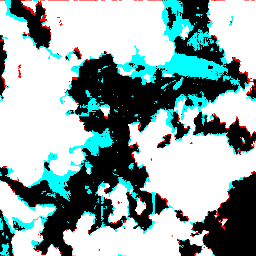}}\\
    \adjustbox{width=\linewidth, frame=0pt 1pt}{\includegraphics{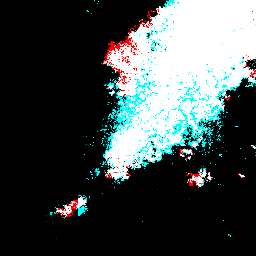}}
    \end{subfigure}
\begin{subfigure}{0.1\textwidth}
    \caption*{PSPNet}
    \adjustbox{width=\linewidth, frame=0pt 1pt}{\includegraphics{other_visuals/sparc_comparison1/pspnet_LC80010812013365LGN00_cloud_mask_RGB.png}}\\
    \adjustbox{width=\linewidth, frame=0pt 1pt}{\includegraphics
    {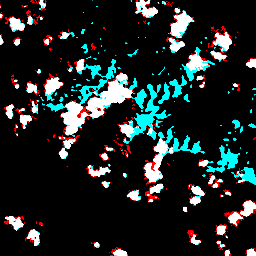}}\\
    \adjustbox{width=\linewidth, frame=0pt 1pt}{\includegraphics{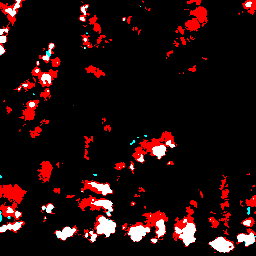}}\\
    \adjustbox{width=\linewidth, frame=0pt 1pt}{\includegraphics{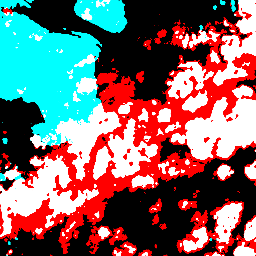}}\\
    \adjustbox{width=\linewidth, frame=0pt 1pt}{\includegraphics{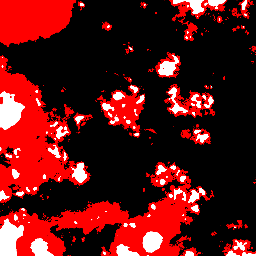}}\\
    \adjustbox{width=\linewidth, frame=0pt 1pt}{\includegraphics{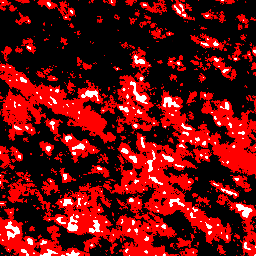}}\\
    \adjustbox{width=\linewidth, frame=0pt 1pt}{\includegraphics{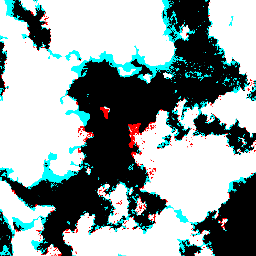}}\\
    \adjustbox{width=\linewidth, frame=0pt 1pt}{\includegraphics{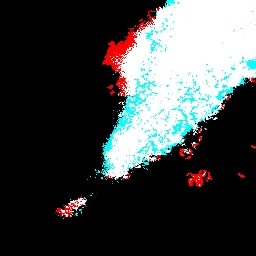}}
    \end{subfigure}
\begin{subfigure}{0.1\textwidth}
    \caption*{RS-Net}
    \adjustbox{width=\linewidth, frame=0pt 1pt}{\includegraphics{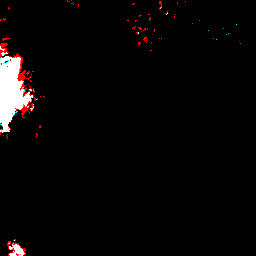}}\\
    \adjustbox{width=\linewidth, frame=0pt 1pt}{\includegraphics
    {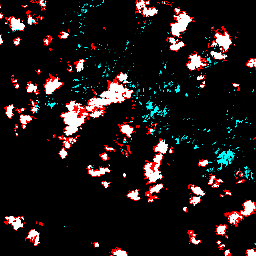}}\\
    \adjustbox{width=\linewidth, frame=0pt 1pt}{\includegraphics{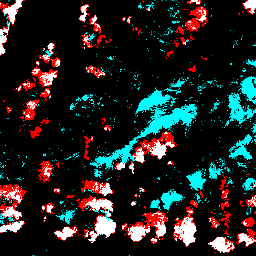}}\\
    \adjustbox{width=\linewidth, frame=0pt 1pt}{\includegraphics{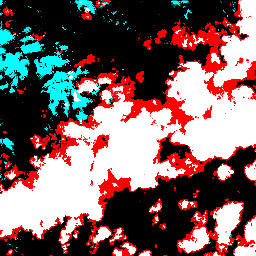}}\\
    \adjustbox{width=\linewidth, frame=0pt 1pt}{\includegraphics{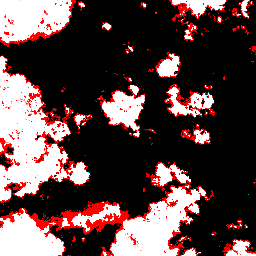}}\\
    \adjustbox{width=\linewidth, frame=0pt 1pt}{\includegraphics{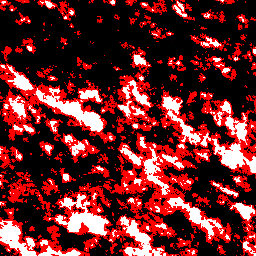}}\\
    \adjustbox{width=\linewidth, frame=0pt 1pt}{\includegraphics{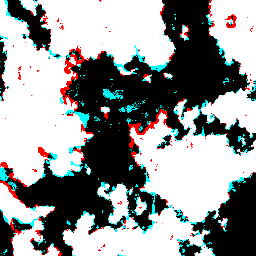}}\\
    \adjustbox{width=\linewidth, frame=0pt 1pt}{\includegraphics{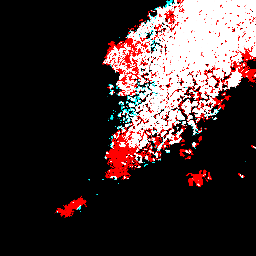}}
    \end{subfigure}
\begin{subfigure}{0.1\textwidth}
    \caption*{RefUNetV4}
    \adjustbox{width=\linewidth, frame=0pt 1pt}{\includegraphics{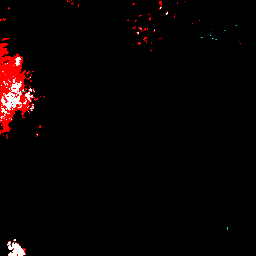}}\\
    \adjustbox{width=\linewidth, frame=0pt 1pt}{\includegraphics
    {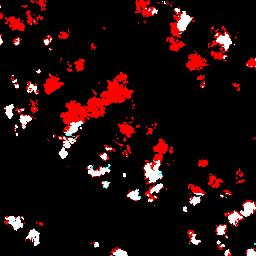}}\\
    \adjustbox{width=\linewidth, frame=0pt 1pt}{\includegraphics{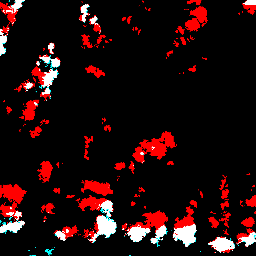}}\\
    \adjustbox{width=\linewidth, frame=0pt 1pt}{\includegraphics{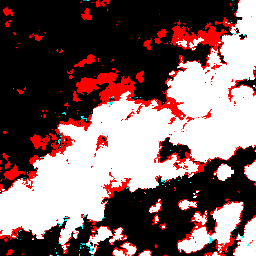}}\\
    \adjustbox{width=\linewidth, frame=0pt 1pt}{\includegraphics{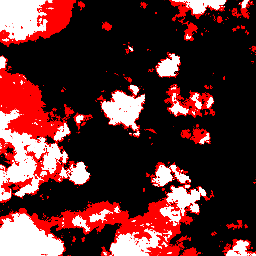}}\\
    \adjustbox{width=\linewidth, frame=0pt 1pt}{\includegraphics{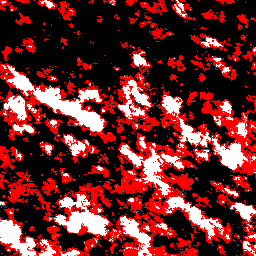}}\\
    \adjustbox{width=\linewidth, frame=0pt 1pt}{\includegraphics{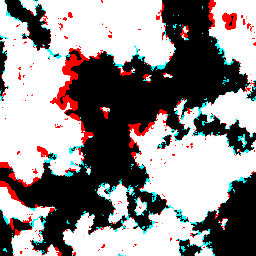}}\\
    \adjustbox{width=\linewidth, frame=0pt 1pt}{\includegraphics{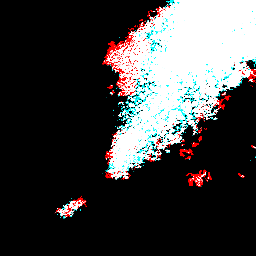}}
    \end{subfigure}
\begin{subfigure}{0.1\textwidth}
    \caption*{CLiSA (ours)}
    \adjustbox{width=\linewidth, frame=0pt 1pt}{\includegraphics{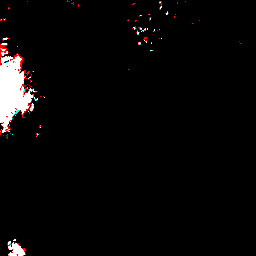}}\\
    \adjustbox{width=\linewidth, frame=0pt 1pt}{\includegraphics
    {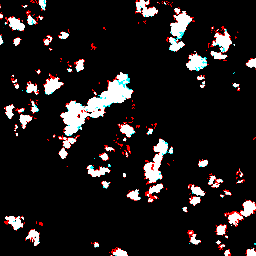}}\\
    \adjustbox{width=\linewidth, frame=0pt 1pt}{\includegraphics{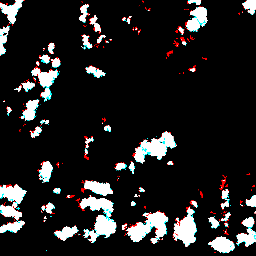}}\\
    \adjustbox{width=\linewidth, frame=0pt 1pt}{\includegraphics{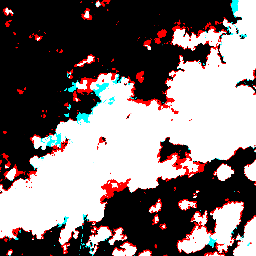}}\\
    \adjustbox{width=\linewidth, frame=0pt 1pt}{\includegraphics{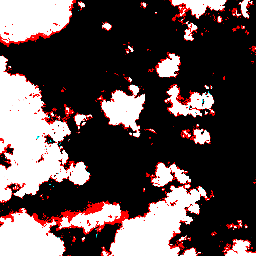}}\\
    \adjustbox{width=\linewidth, frame=0pt 1pt}{\includegraphics{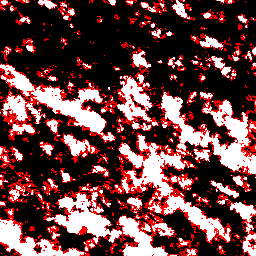}}\\
    \adjustbox{width=\linewidth, frame=0pt 1pt}{\includegraphics{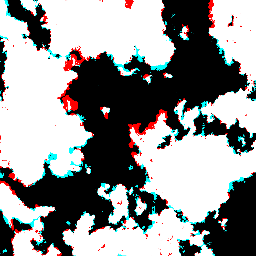}}\\
    \adjustbox{width=\linewidth, frame=0pt 1pt}{\includegraphics{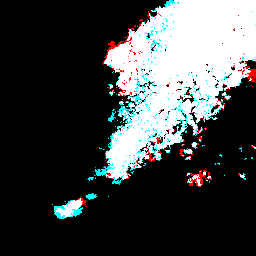}}
    \end{subfigure}
Legend \quad \tikz \fill [red] rectangle (1, 0.2); Omission error \quad
\tikz \fill [Mycolor2] rectangle (1, 0.2); Commission error
\caption{Comparison of cloud mask generation results on different datasets. (a)-(c): SPARCS dataset, (d)-(f): Sentinel-2 dataset, (g)-(h): Cartosat-2s dataset. }
\label{fig:fig2_results}
\end{figure*}

\subsubsection{Qualitative analysis}
Figure \ref{fig:fig1_results} shows cloud segmentation visual outcomes on the Biome dataset with different signatures. For effective visual assessment, we denote the cloudy and non-cloudy pixels in white and black as well as mark the misinterpreted cloudy and non-cloudy ones in red and cyan, respectively. As shown in the figure, our proposed method generates a cloud mask closest to the reference compared to baseline techniques, especially for the signatures with urban, snow, and water bodies. Although not designed for remote sensing applications, PSPnet \cite{22}, and, DeeplabV3+ \cite{23} results in substantially accurate cloud mask in most of the cases except in situations like thin cloud and snow-cloud coexistence. In these situations, they lead to commission and commission errors, respectively. RefUNetV4 \cite{4010} can generate a cloud mask very close to the reference, however, it still suffers from commission and commission errors in scenarios like snow and urban regions. CDNetV2 \cite{19} on the other hand failed miserably in these situations by giving misclassified results most of the time as shown in the figure. Compared to that CDNetV1 \cite{19} and RS-Net \cite{18} are able to generate reasonably accurate cloud masks. However, as their performance results in low Kappa in Table \ref{tab:my-table}, their performance is situational and they also provide misclassified outcomes. Our method, in contrast, performs preferably by generating accurate cloud mask confidently in most of the cases along with the mentioned critical cases which also supports our quantitative analysis in Table \ref{tab:my-table}.
\subsection{Analysis on Sparsac and other datasets}
\begin{figure}
    \centering
    \begin{subfigure}{0.465\linewidth}
        \includegraphics[width=\linewidth]{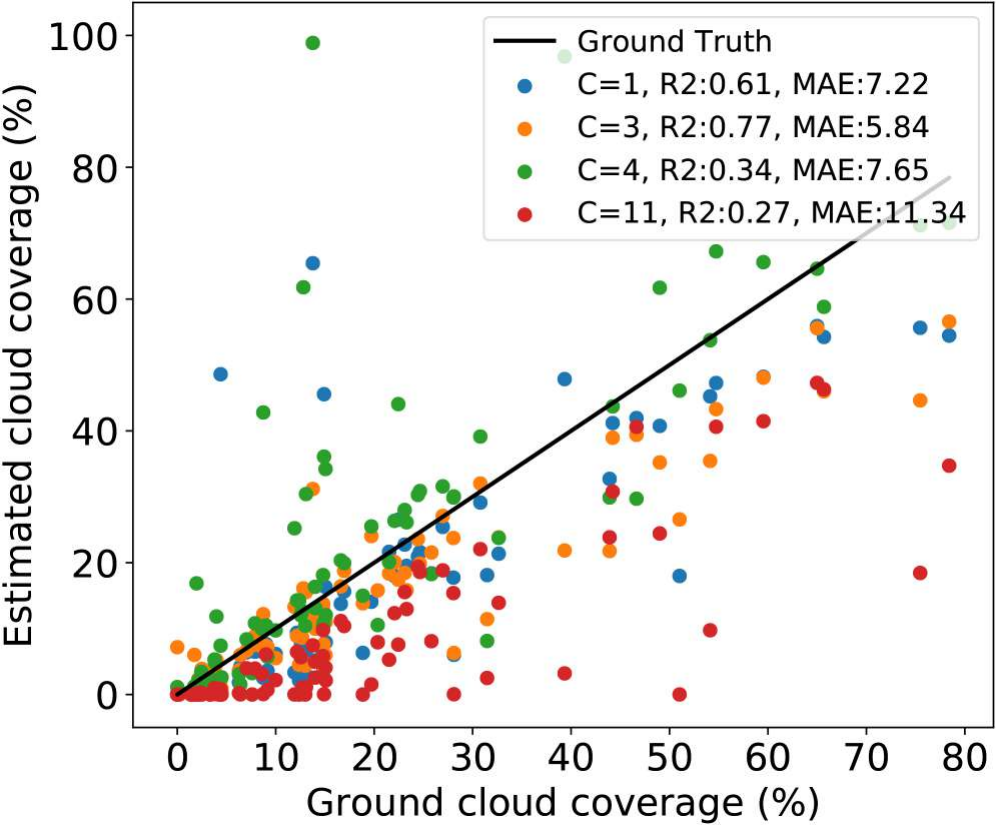}
        \caption{}
    \end{subfigure}
    \begin{subfigure}{0.465\linewidth}
        \includegraphics[width=\linewidth]{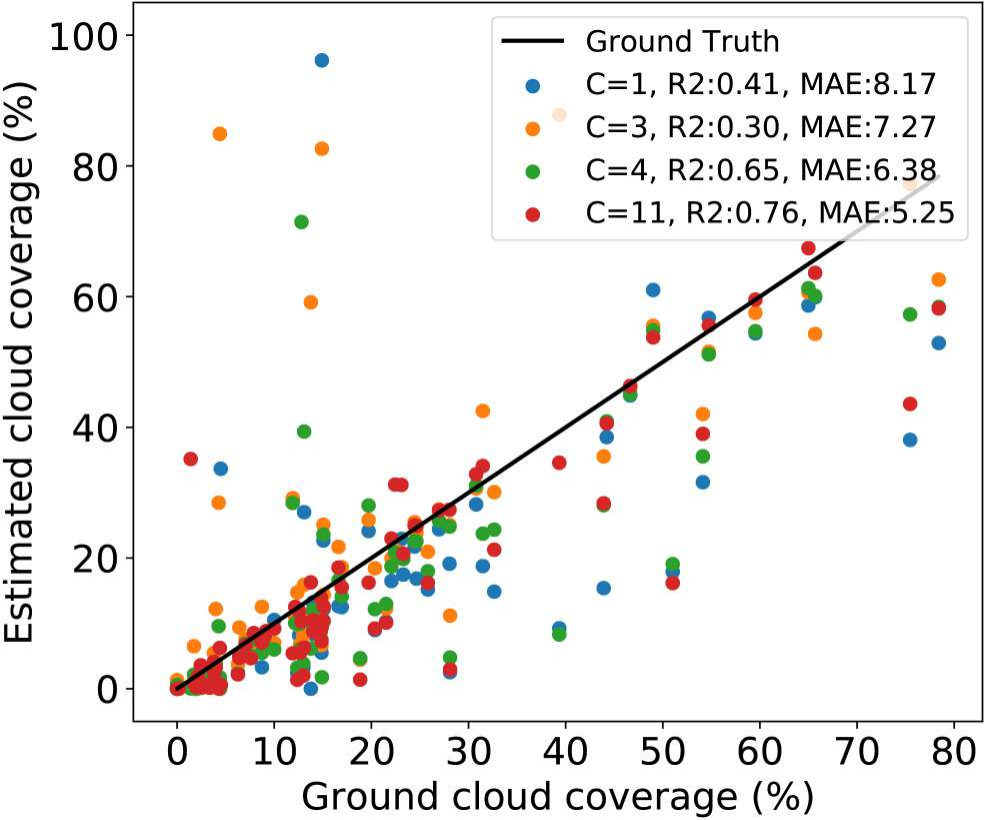}
        \caption{}
    \end{subfigure}
    \begin{subfigure}{0.465\linewidth}
        \includegraphics[width=\linewidth]{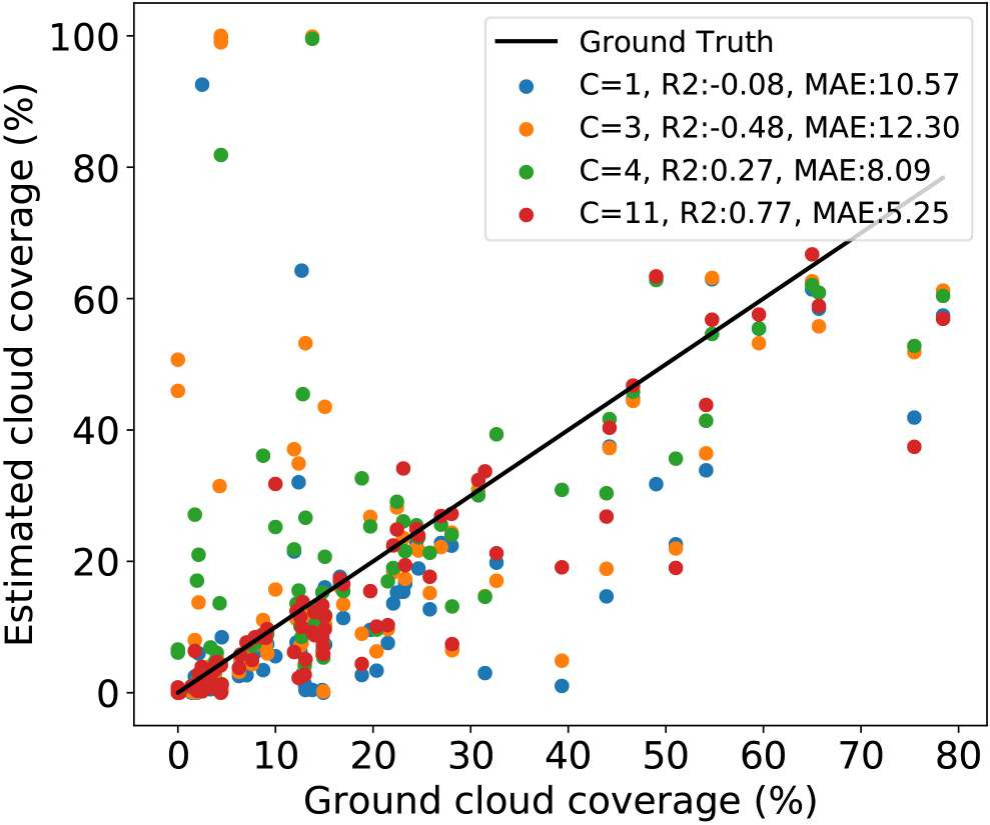}
        \caption{}
    \end{subfigure}
    \begin{subfigure}{0.465\linewidth}
        \includegraphics[width=\linewidth]{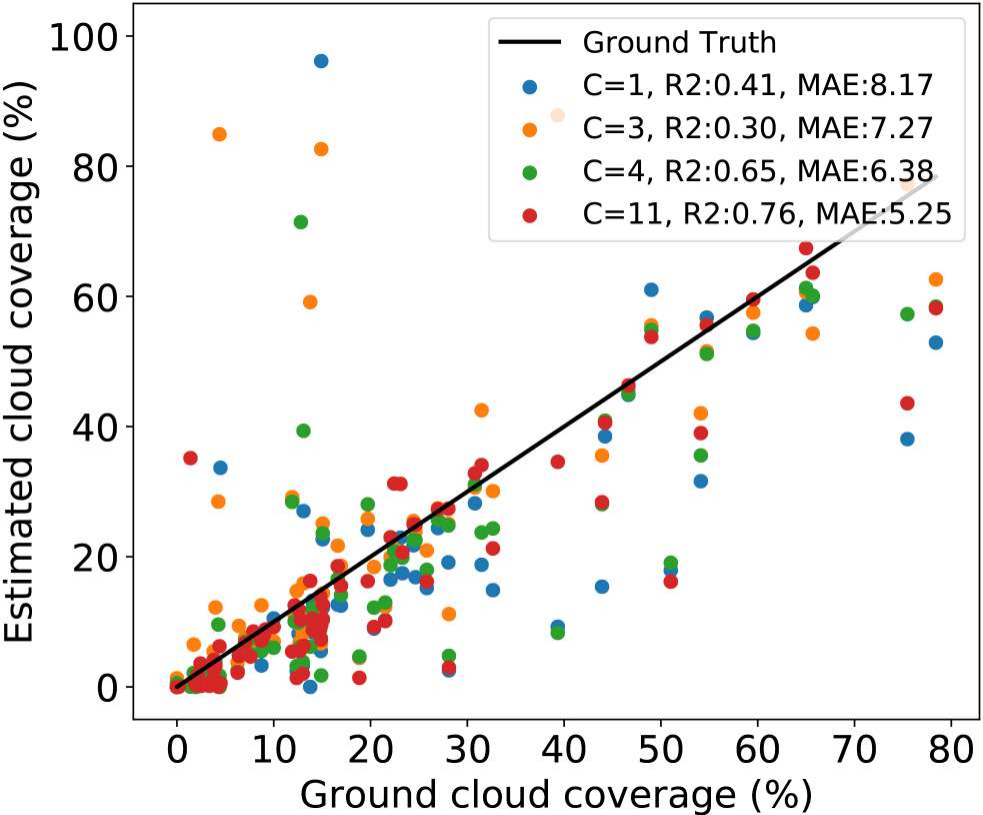}
        \caption{}
    \end{subfigure}
    \begin{subfigure}{0.465\linewidth}
        \includegraphics[width=\linewidth]{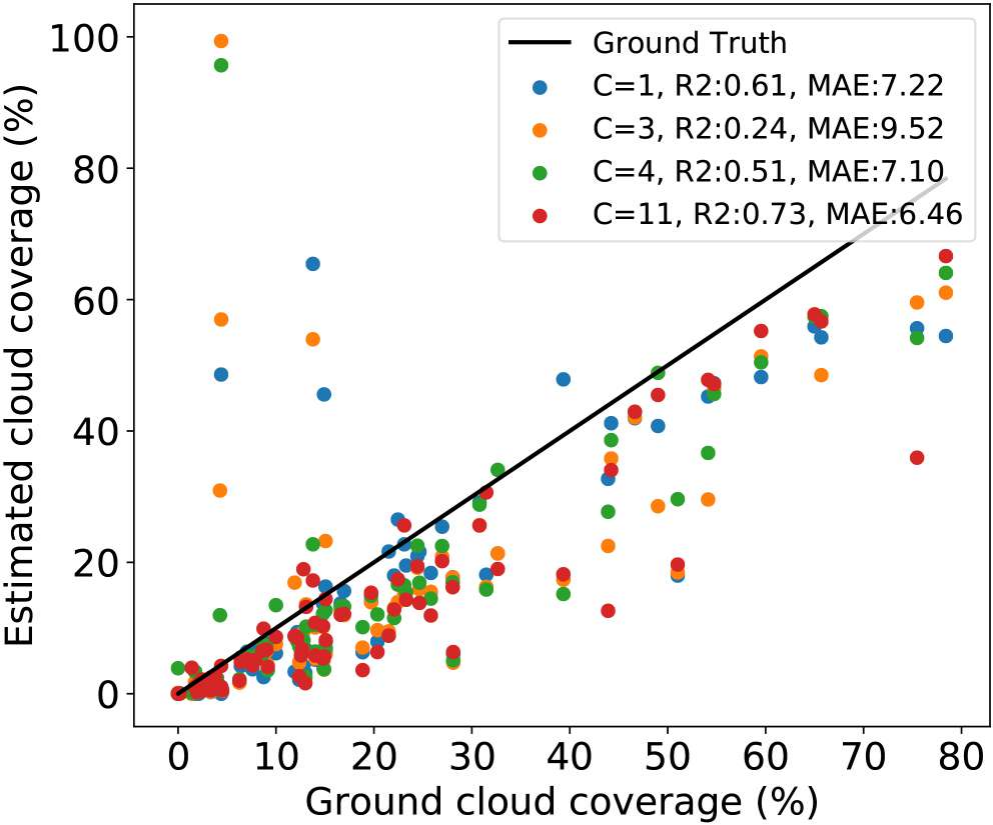}
        \caption{}
    \end{subfigure}
    \begin{subfigure}{0.465\linewidth}
        \includegraphics[width=\linewidth]{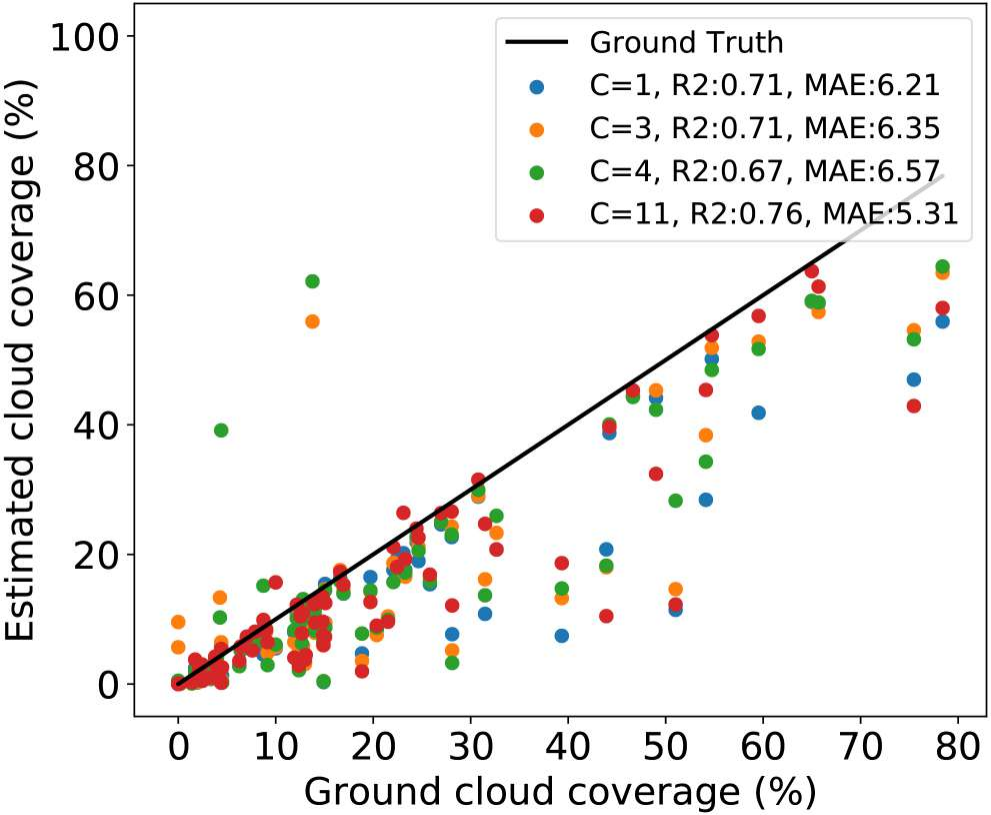}
        \caption{}
    \end{subfigure}
    \begin{subfigure}{0.465\linewidth}
        \includegraphics[width=\linewidth]{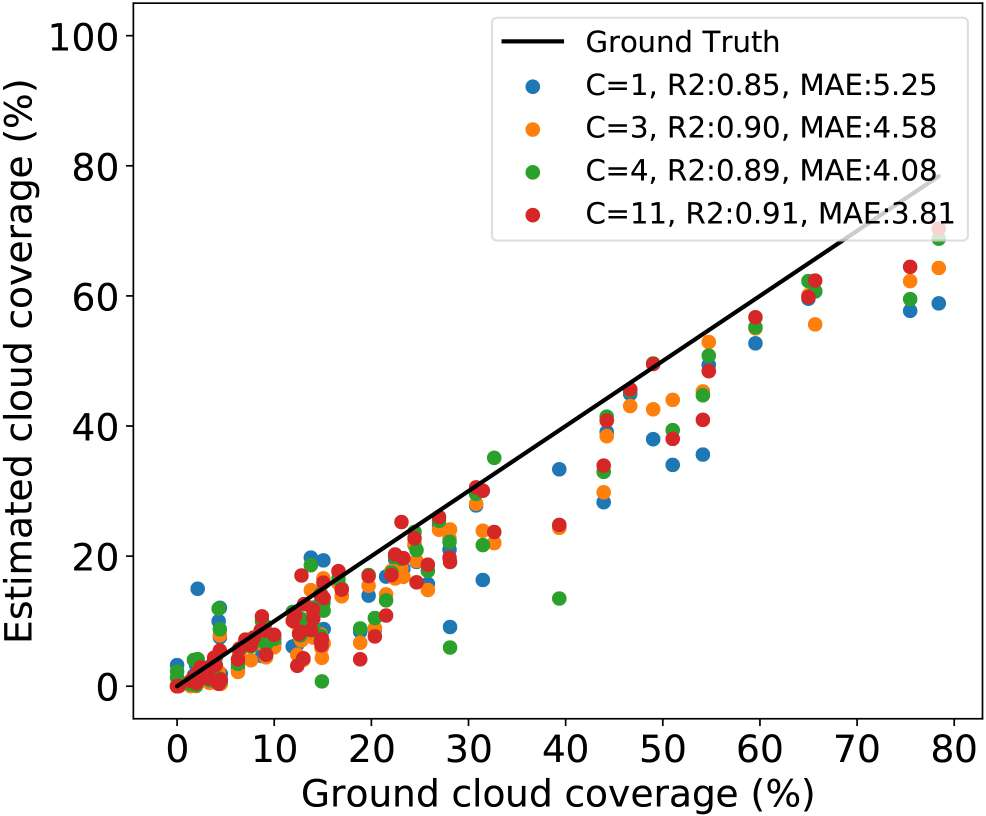}
        \caption{}
    \end{subfigure}
    \caption{Comparison cloud coverage estimation on SPARCS dataset among (a) CDNetv2, (b) CDNetv1, (c) DeepLabV3+, (d) PSPnet, (e) RS-Net, (f) RefUNetV4, and, (g) CLiSA (ours).}
    \label{fig:scatter_plot}
\end{figure}
\subsubsection{Quantitative analysis}
Table \ref{tab:my-table2} illustrates quantitative results on the SPARCS dataset. Similar to the previous case, our method results in the best comprehensive outcomes for all band combinations. Compared to the second best method, our model achieves 1-3\% improvement in OA, 2-5\% in mIoU, 1-5\% in Kappa, 70-80\% in BI, and 35-50\% in HD. This observation is crucial as these results are generated using the weights corresponding to the training of the Biome dataset. Hence, this represents a clear picture of the generalization capability for each module. For better visualization, we have further estimated cloud coverage for each model and evaluated its accuracy with respect to the ground truth as shown in Figure \ref{fig:scatter_plot}. From the figure, our estimated cloud coverage results are more closely distributed to the benchmark line for all 4 band combinations compared to others. This also can be depicted from the R2-scores and MAE (mean absolute error) values recorded in Figure \ref{fig:scatter_plot}. Our method achieves the highest R2-score and lowest MAE for all band combinations $(C=\{1,3,4,11\})$ followed by RefUNetV4. All other methods mostly perform substantially for particular band combinations as outlined in the scatter plots. This demonstrates that our method is more robust for any band configuration and has better generalization capability compared to other state-of-the-art methods.\par
Table \ref{tab:my-table3} and \ref{tab:my-table4} refers to our quantitative study of cloud mask generation for Sentinel-2 and Cartosat-2S dataset, respectively. In both cases, the results are generated for band combination $C=\{1,3,4\}$ using transfer learning for 2 iterations. In the first case, when our proposed method achieves a performance gain of 5-7\% in OA, 6-8\% in mIoU, 6-7\% in Kappa, 10-23\% in BI, and, 13-20\% in HD, for the latter case, it is 2-4\% improvement in OA, 3-5\% in mIoU, 8-20\% in BI, and, 47-60\% in HD with respect to corresponding second best scenario. From this observation, we can say for cross-domain cloud mask generation, our model has the best generalization potential and can perform most robustly with minimal computation.

\begin{table}[]
\caption{Quantitative ablation study for introducing different loss functions}
\label{tab:my-table5}
\resizebox{\columnwidth}{!}{%
\begin{tabular}{cccc|ccccccc}
\hline
\begin{tabular}[c]{@{}c@{}}Focal+L2\\ Loss\end{tabular} & \begin{tabular}[c]{@{}c@{}}Jaccard\\ Loss\end{tabular} & \begin{tabular}[c]{@{}c@{}}Lovasz\\ Loss\end{tabular} & \begin{tabular}[c]{@{}c@{}}Adversarial\\ Loss\end{tabular} & OA             & Rec            & Pre            & mIoU(\%)       & Kappa          & BI(\%)         & HD (m)          \\ \hline
                                                \cmark        &  \xmark                                                      &    \xmark                                                   &   \xmark                                                         & 93.97          & 93.58          & 94.35          & 86.88          & 88.49          & 42.29          & 923.57          \\
   \cmark                                                     &   \cmark                                                     &    \xmark                                                   &      \xmark                                                     & 95.29          & 93.98          & 93.76          & 88.06          & 89.06          & 45.59          & 952.18          \\
                                        \cmark                &       \xmark                                                 & \cmark                                                      &           \xmark                                                 & 97.12          & 96.44          & 94.73          & 94.25          & 90.26          & 55.71          & 682.15          \\
                                       \cmark                 &   \xmark                                                     &  \cmark                                                     &     \cmark                                                       & \textbf{98.05} & \textbf{97.78} & \textbf{96.88} & \textbf{96.02} & \textbf{94.09} & \textbf{59.33} & \textbf{497.12} \\ \hline
\end{tabular}%
}
\vspace{-0.25cm}
\end{table}
\begin{table}[]
\centering
\caption{Quantitative ablation study for introducing different architectural elements}
\label{tab:my-table6}
\resizebox{\columnwidth}{!}{%
\begin{tabular}{ccc|ccccccc}
\hline
\begin{tabular}[c]{@{}c@{}}Without\\ Attention\end{tabular} & \begin{tabular}[c]{@{}c@{}}With\\ DOSA\end{tabular} & \begin{tabular}[c]{@{}c@{}}With\\ DOSA+HC2A\end{tabular} & OA             & Rec            & Pre            & mIoU(\%)       & Kappa          & BI(\%)         & HD (m)          \\ \hline
                                                     \cmark       &       \xmark                                              & \xmark                                                          & 94.09          & 96.41          & 92.88          & 92.25          & 90.11          & 48.45          & 742.29          \\
                                                    \xmark        &      \cmark                                               &  \xmark                                                        & 95.26          & 94.87          & 94.02          & 93.73          & 92.19          & 53.37          & 662.28          \\
                                                 \xmark           &      \cmark                                               &    \cmark                                                      & \textbf{98.05} & \textbf{97.78} & \textbf{96.88} & \textbf{96.02} & \textbf{94.09} & \textbf{59.33} & \textbf{497.12} \\ \hline
\end{tabular}%
}
\vspace{-0.25cm}
\end{table}

\subsubsection{Qualitative analysis}
Figure \ref{fig:fig2_results} illustrates visual results of cloud mask generation for three datasets, SPARCS, Sentinel-2, and Cartosat-2S. Similar to the Biome scenario, our proposed method outperforms all other benchmark methods for all the datasets. CDNetV2 in all aspects unable to perform well, especially for complex scenarios like snow-cloud co-existence and thin clouds. For all the datasets, it fails to differentiate between snow and cloud, and to precisely estimate thin clouds. Compared to that, CDNetV1 and DeepLabV3+ are able to distinguish between snow and cloud features for the SPARSAC dataset. However, they are unable to perform the same for the other two datasets, and they substantially suffer from commission errors. The performance of PSPnet and RS-Net is situational, and for most of the complicated cases, they also result in misclassified output. On the other hand, RefUNetV4 performs efficiently in these situations. However, it still suffers from cases of commission error. Our proposed model is able to precisely extract cloud masks from different band combinations with minimal error in all situations and also follows the boundary of the segmented masks very closely compared to the other methods. This also supports our quantitative analysis in the above Tables. Next section, we discuss the ablation study.
    
    
\begin{figure*}[]
\captionsetup[subfigure]{justification=centering}
\centering
\begin{subfigure}{0.13\textwidth}
    \caption*{Input Image} 
    \adjustbox{width=\linewidth, frame=0pt 1pt}{\includegraphics{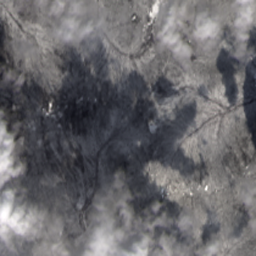}}\\
    \adjustbox{width=\linewidth, frame=0pt 1pt}{\includegraphics{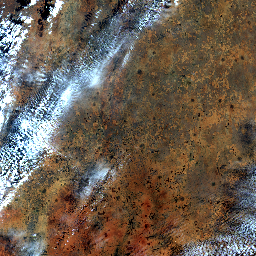}}
    \end{subfigure}   
\begin{subfigure}{0.13\textwidth}
    \caption*{Ground Truth}
    \adjustbox{width=\linewidth, frame=0pt 1pt}{\includegraphics{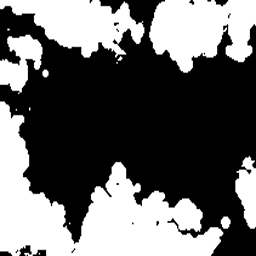}}\\
    \adjustbox{width=\linewidth, frame=0pt 1pt}{\includegraphics{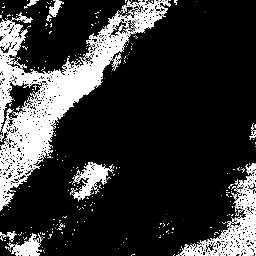}}
    \end{subfigure}
\begin{subfigure}{0.13\textwidth}
    \caption*{Focal+L2}
    \adjustbox{width=\linewidth, frame=0pt 1pt}{\includegraphics{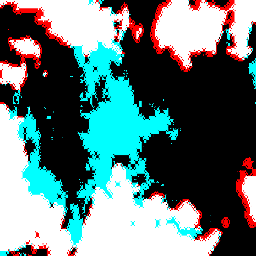}}\\
    \adjustbox{width=\linewidth, frame=0pt 1pt}{\includegraphics{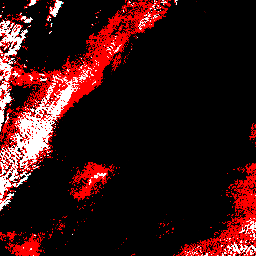}}
    \end{subfigure}   
\begin{subfigure}{0.13\textwidth}
    \caption*{Focal+L2+Jacard}
    \adjustbox{width=\linewidth, frame=0pt 1pt}{\includegraphics{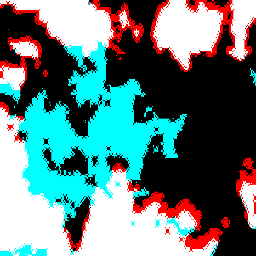}}\\
    \adjustbox{width=\linewidth, frame=0pt 1pt}{\includegraphics{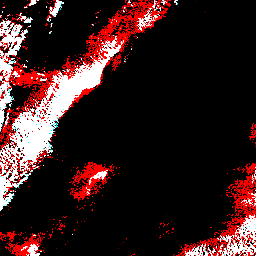}}
    \end{subfigure}    
\begin{subfigure}{0.13\textwidth}
    \caption*{Focal+L2+Lovasz}
    \adjustbox{width=\linewidth, frame=0pt 1pt}{\includegraphics{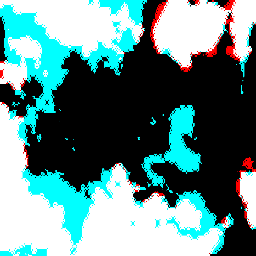}}\\
    \adjustbox{width=\linewidth, frame=0pt 1pt}{\includegraphics{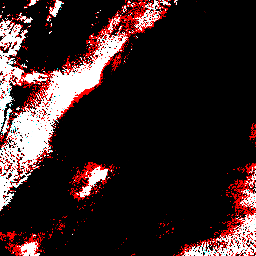}}
    \end{subfigure}
\begin{subfigure}{0.13\textwidth}
    \caption*{Focal+L2+Lovasz\\+Adversarial}
    \adjustbox{width=\linewidth, frame=0pt 1pt}{\includegraphics{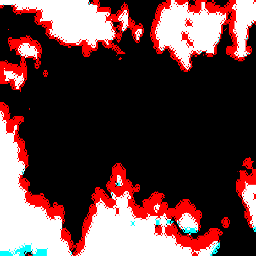}}\\
    \adjustbox{width=\linewidth, frame=0pt 1pt}{\includegraphics{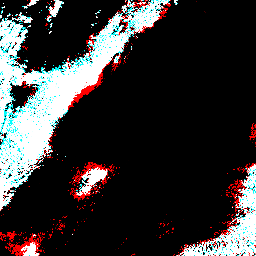}}
    \end{subfigure}\par
    Legend \quad \tikz \fill [red] rectangle (1, 0.2); Omission error \quad
\tikz \fill [Mycolor2] rectangle (1, 0.2); Commission error
    \caption{Visual comparison of introducing different loss functions}
    \label{fig:loss_abl}
\end{figure*}
\begin{figure*}[!ht]
\centering
\begin{subfigure}{0.15\linewidth}
    \caption*{Input Image} 
    \adjustbox{width=\linewidth, frame=0pt 1pt}{\includegraphics{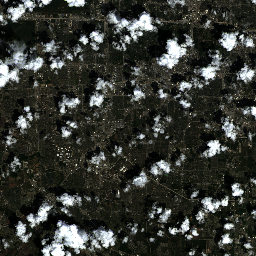}}\\
    \adjustbox{width=\linewidth, frame=0pt 1pt}{\includegraphics{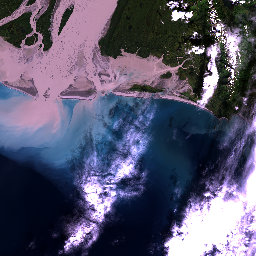}}
    \end{subfigure}   
\begin{subfigure}{0.15\textwidth}
    \caption*{Ground Truth}
    \adjustbox{width=\linewidth, frame=0pt 1pt}{\includegraphics{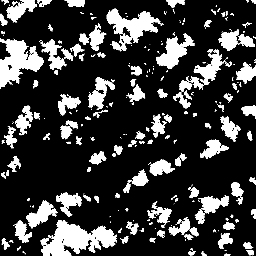}}\\
    \adjustbox{width=\linewidth, frame=0pt 1pt}{\includegraphics{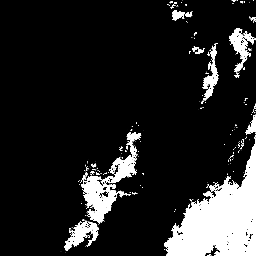}}
    \end{subfigure}
\begin{subfigure}{0.15\textwidth}
    \caption*{Without attention}
    \adjustbox{width=\linewidth, frame=0pt 1pt}{\includegraphics{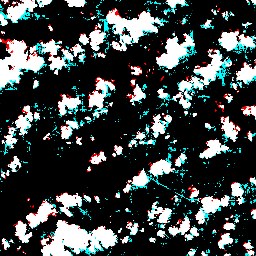}}\\ 
    \adjustbox{width=\linewidth, frame=0pt 1pt}{\includegraphics{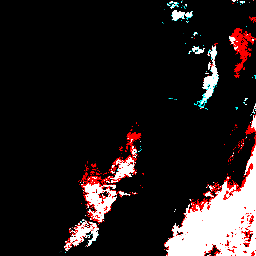}}
    \end{subfigure}   
\begin{subfigure}{0.15\textwidth}
    \caption*{After DOSA}
    \adjustbox{width=\linewidth, frame=0pt 1pt}{\includegraphics{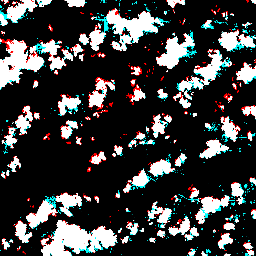}}\\
    \adjustbox{width=\linewidth, frame=0pt 1pt}{\includegraphics{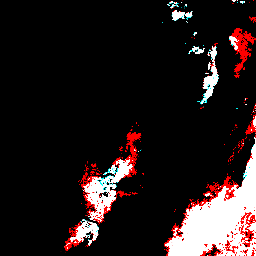}}
    \end{subfigure}    
\begin{subfigure}{0.15\textwidth}
    \caption*{After DOSA+HC2A}
    \adjustbox{width=\linewidth, frame=0pt 1pt}{\includegraphics{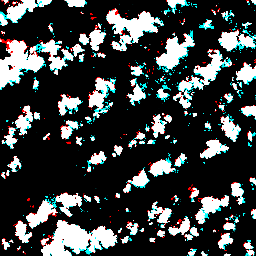}}\\
    \adjustbox{width=\linewidth, frame=0pt 1pt}{\includegraphics{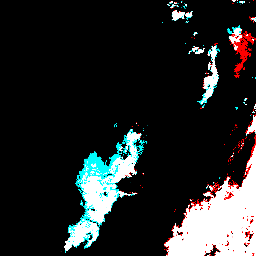}}
    \end{subfigure}\par
    Legend \quad \tikz \fill [red] rectangle (1, 0.2); Omission error \quad
\tikz \fill [Mycolor2] rectangle (1, 0.2); Commission error
    \caption{Visual comparison of introducing different architectural components}
    \label{fig:arch_abl}
\end{figure*}
\section{Ablation Study}\label{abl}
In this section, we perform ablation studies related to investigating the efficacy of different loss functions and architectural elements like DOSA and HC2A along with their complexity analysis and also discuss about their overall Lipschitz stability and adversarial robustness.
\subsection{Choice of objective functions}
Table \ref{tab:my-table5} shows quantitatively the effect of each loss function on the performance of our model for the Biome dataset with $C=11$ band configuration. As discussed in section \ref{obj}, introducing Lov\'asz-Softmax Loss instead of traditional Jaccard loss improves the overall performance with almost 2\% gain in OA, more than 6\% in mIoU and 10\% in BI. These results are further enhanced by introducing adversarial loss which achieved a performance gain of 4\% in Kappa and 27\% in HD. Figure \ref{fig:loss_abl} visualizes the effect of each loss for estimating cloud mask. We can see that due to introducing Lov\'asz loss, our model can generate a cloud mask with better boundary accuracy and fewer misinterpreted cases. However, by applying adversarial loss, our model estimates precise cloud masks more confidently as we are forcing the model to mimic the reference manifold with this objective function.
\subsection{Choice of Architectural components}
Here, we study the significance of two proposed modules, DOSA, and HC2A in generating precise cloud masks. Table \ref{tab:my-table6} illustrates the comprehensive performance of our model on the Biome dataset for $C=11$ in the presence of proposed attention modules. From the table, it is clear that introducing DOSA and HC2A improves in performance of our model by 4\% in OA, mIoU, and, Kappa, 11\% in BI, and, 33\% in HD with respect to the baseline model. Figure \ref{fig:arch_abl} shows the effect of these attention modules in cloud segmentation visually. Without attention modules, the model results in commission errors. The introduction of DOSA mitigates this issue up to a certain extent, however, it also increases commission error as depicted in the figure. Presenting HC2A along with DOSA, not only reduces the overall misclassifications it also improves the generalization capability of our model.
\begin{figure}[]
\captionsetup[subfigure]{justification=centering}
    \centering
\begin{subfigure}{0.3\linewidth}
    \caption*{Feature map ($X_1$) \\ without attention} 
    \includegraphics[width = \linewidth]{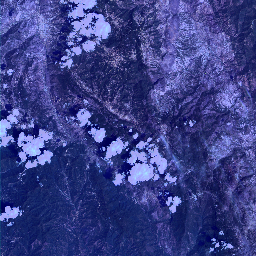}\\
\end{subfigure}
\begin{subfigure}{0.3\linewidth}
    \caption*{Feature map ($Z_1$) \\after DOSA} 
    \includegraphics[width = \linewidth]{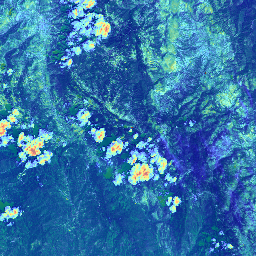}\\
\end{subfigure}
\begin{subfigure}{0.3\linewidth}
    \caption*{Feature Map ($Y_1$) \\after DOSA+HC2A} 
    \includegraphics[width = \linewidth]{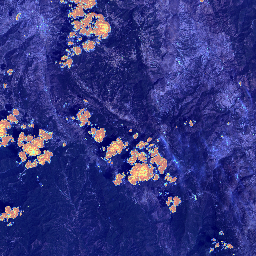}\\
\end{subfigure}   
    \caption{Visualization of effect of each attention modules on feature map}
    \label{fig:feature_map}
\end{figure}
It can be visualized from the second example of Figure \ref{fig:arch_abl}, where there is an commission error in reference to the positions of thin clouds. However, using both these attention modules, our model can faithfully mark these cases as cloudy pixels. The main reason behind these can be visualized in Figure \ref{fig:feature_map} where we present an illustration of the feature map after each attention module. It is readily understandable that without any attention, the model prioritizes every feature equally. Whereas applying only DOSA, focuses more on cloudy regions, however, it also focuses on redundant features. After employing DOSA+HC2A, it not only gives more emphasis to the cloudy pixels but also suppresses the surrounding redundant features.
\begin{figure}[]
\captionsetup[subfigure]{justification=centering}
    \centering
\begin{subfigure}{0.45\linewidth}
    \caption*{Robustness analysis under \\ FGSM attack} 
    \includegraphics[height=3cm, width = \linewidth]{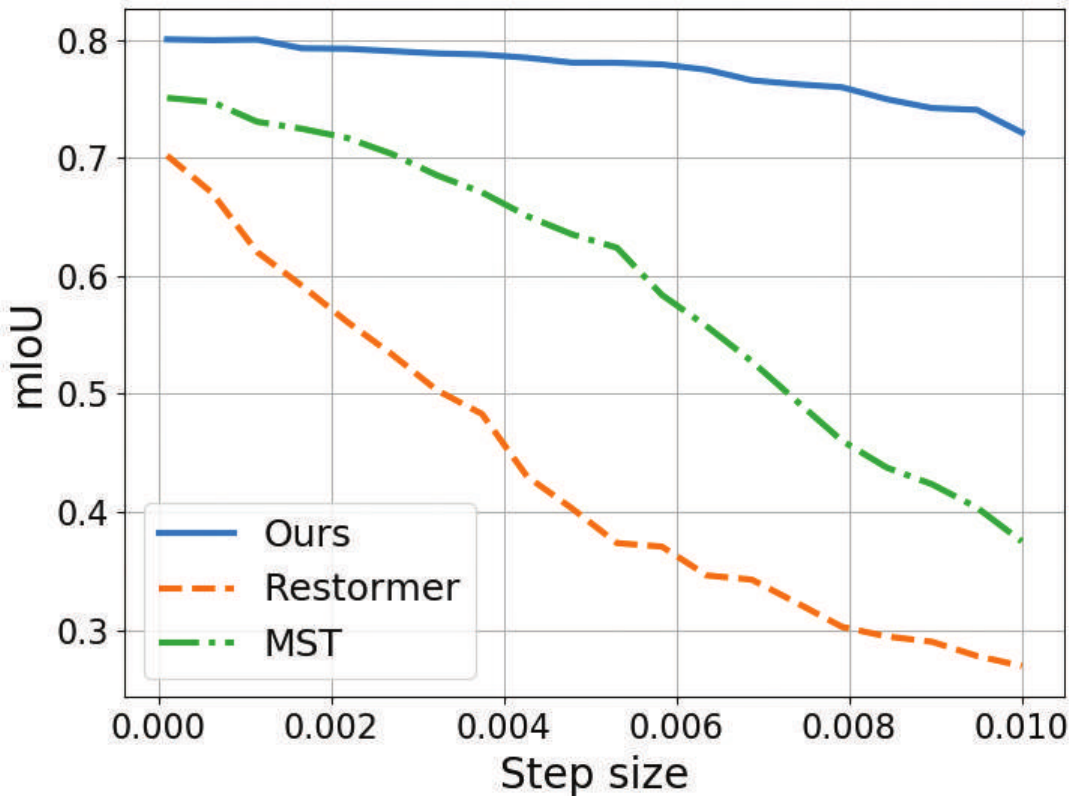}
\end{subfigure}
\begin{subfigure}{0.45\linewidth}
    \caption*{Robustness analysis under \\ PGD-20 attack} 
    \includegraphics[height=3cm, width = \linewidth]{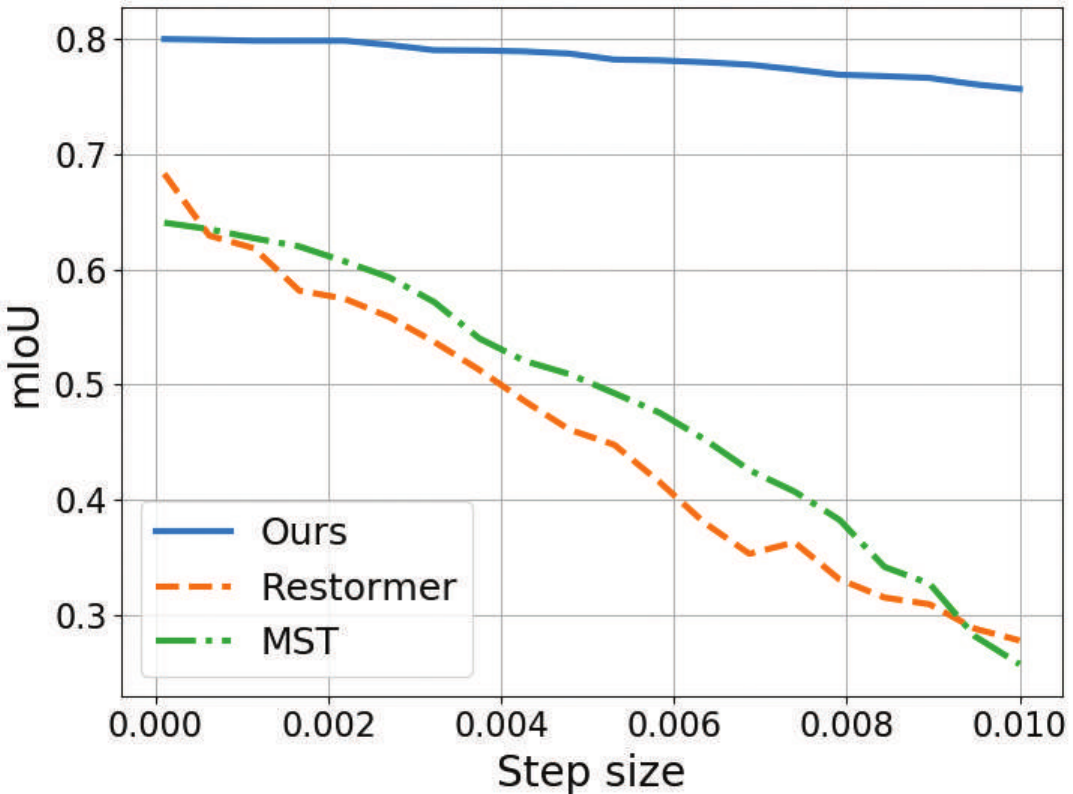}
\end{subfigure}   
    \caption{Analysis of adversarial robustness against FGSM and PGD-20}
    \label{fig:adv_atk}
\end{figure}
\subsection{Complexity analysis}
Here we analyze the complexity of our proposed attention modules for an input feature $X$ of shape $H\times W \times C$. In the case of DOSA, the computation of attention scores in both spatial and channel attention branches involves, (1) matrix multiplication between key and value arrays which results in computational complexity of $\mathcal{O}(HWC)$ and (2) element-wise multiplication with value with complexity $\mathcal{O}(HWC)$. Therefore, the overall complexity for DOSA is $\mathcal{O}(HWC)$, which makes it a linear self-attention. The estimation of attention in HC2A involves matrix multiplication between key and value arrays with complexity $\mathcal{O}(C^2)$, followed by matrix multiplication with value of shape $HW\times C$. Therefore, overall memory complexity of HC2A is $\mathcal{O}(HWC^2)$, and for combination of DOSA+HC2A, it will be $\mathcal{O}(HWC+HWC^2)$.
\subsection{Lipschitz stability and adversarial robustness}
To validate our derived \textbf{Theorem 1} and \textbf{Theorem 2}, we empirically estimate the 2-Lipschitz constant by perturbing specific input channel (denoted as $i$ in Table \ref{tab:my-table7}) of the particular feature map. The diagonal elements Jacobian corresponding to computing this Lipschitz coefficient is estimated by observing the related change in the output of the attention modules based on the perturbed input. Table \ref{tab:my-table7}, shows the estimated Lipschitz constant for our DOSA+HC2A module and is also compared with that of multihead-self-attention transformer (MST). As shown in the table, the spectral norm of diagonal elements of the Jacobian corresponding to different perturbed channels is comparatively less for our proposed attention modules with respect to that of MST. This not only supports our argument from the derived Theorems but also suggests that our proposed method has better stability in terms of generating robust output in the presence of any perturbed input. Due to this enhanced Lipschitz stability, it also implies robustness against adversarial attacks as studied in \cite{4004}. To validate this, we perform experiments of our model against the Fast Gradient Sign Method (FGSM) and Projected Gradient Descent (PGD) attacks and compare the performance with MST and Restormer. For all the cases, we perturb the input MX image by tuning step size for both FGSM and PGD-20 attacks. As shown in Figure \ref{fig:adv_atk}, our method results in more stable performance in terms of mIoU compared to other self-attention modules as the perturbation to the input increases for both attacks. therefore, this supports our argument of providing better stability by introducing the DOSA+HC2A module which also provides better robustness against adversarial attacks.
\begin{table}[]
\caption{Some of empirically estimated 2-Lipschitz constant of proposed attention modules for different perturbed channels}
\label{tab:my-table7}
\begin{tabular}{c|c|c|c}
\hline
$i$  & \begin{tabular}[c]{@{}c@{}} $||J_{i,i}||_2$ \\ for MST\end{tabular} & \begin{tabular}[c]{@{}c@{}}$||J_{i,i}||_2$ \\ for DOSA\end{tabular} & \begin{tabular}[c]{@{}c@{}}$||J_{i,i}||_2$ \\ for DOSA+HC2A\end{tabular} \\ \hline
0  & 2.32                                                   & 0.86                                                    & 0.69                                                         \\
5  & 1.59                                                   & 0.94                                                    & 0.74                                                         \\
25 & 1.88                                                   & 1.26                                                    & 0.78                                                         \\
50 & 0.78                                                   & 0.73                                                    & 0.62                                                         \\ \hline
\end{tabular}%
\end{table}
\section{Conclusion}\label{conc}
In this study, we propose a novel hybrid transformer architecture, namely CLiSA - cloud segmentation via Lipschitz stable attention module consisting of a dual orthogonal self-attention (DOSA) and a hierarchical cross-channel attention (HC2A) model. We develop this whole setup under an adversarial setting in the presence of Lov\'asz-Softmax Loss. We both theoretically and empirically derive the Lipschitz stability of our attention module and visualize how it affects in the robustness of the model's performance against various adversarial attacks. The experiment results on multiple satellite datasets like Landsat-8, Sentinel-2, and Cartosat-2S, our model performs reasonably well in extracting accurate cloud masks both qualitatively and quantitatively against other state-of-the-art methods. Our model not only results in the generation of a refined boundary-aware cloud mask but also has better generalization capability. We also undergo an ablation study to analyze how each proposed module affects the outcome.





\ifCLASSOPTIONcaptionsoff
  \newpage
\fi



%


\bibliographystyle{IEEEtran}
\bibliography{main}
%








\end{document}